\documentclass{article}

     \usepackage[preprint]{neurips_2023}

\usepackage[utf8]{inputenc} 
\usepackage[T1]{fontenc}    
\usepackage{hyperref}       
\usepackage{url}            
\usepackage{booktabs}       
\usepackage{amsfonts}       
\usepackage{nicefrac}       
\usepackage{microtype}      
\usepackage{xcolor}         
\usepackage{breakurl}

\usepackage{amsmath}
\usepackage{amssymb}
\usepackage{mathtools}
\usepackage{amsthm}
\usepackage{dsfont}

\usepackage{tikzit}

\tikzstyle{node}=[fill=white, draw=black, shape=circle, minimum size=1mm, ultra thick]
\tikzstyle{small box}=[fill=white, draw=black, shape=rectangle, minimum height=0.5cm, minimum width=0.5cm, ultra thick]
\tikzstyle{weyl}=[fill=white, draw={rgb,255: red,0; green,0; blue,109}, shape=rectangle, minimum height=0.5cm, minimum width=0.5cm]
\tikzstyle{filled_node}=[fill=black, draw=black, shape=circle, minimum size=1mm, ultra thick]
\tikzstyle{large box}=[fill=white, draw=black, shape=rectangle, minimum height=0.5cm, minimum width=1cm, ultra thick]

\tikzstyle{thick}=[-, ultra thick]
\tikzstyle{blue_thick}=[-, ultra thick, draw=blue]
\tikzstyle{dashes}=[-, dashed, draw={rgb,255: red,191; green,191; blue,191}, dash pattern=on 2mm off 1mm, fill={rgb,255: red,244; green,228; blue,0}]
\tikzstyle{thick_arrow}=[ultra thick, ->]
\tikzstyle{dash_1}=[-, dashed]
\tikzstyle{dash_2}=[-, dashed, fill={rgb,255: red,246; green,235; blue,255}]
\tikzstyle{dash_3}=[-, dashed, fill={rgb,255: red,229; green,255; blue,181}]
\tikzstyle{dash_4}=[-, dashed, fill={rgb,255: red,255; green,209; blue,153}]
\tikzstyle{red_thick}=[-, ultra thick, draw=red]
\tikzstyle{dash_5}=[-, dashed, fill={rgb,255: red,225; green,255; blue,254}]
\tikzstyle{jellyfish}=[-, ultra thick, fill={rgb,255: red,34; green,48; blue,255}]

\usepackage{bm}
\usepackage{nicematrix}
\usepackage{comment}
\usepackage{tabularray}

\usepackage{algorithm}
\usepackage{algpseudocode}

\DeclareMathOperator{\Hom}{Hom}

\DeclareMathOperator{\sgn}{sgn}

\theoremstyle{plain}
\newtheorem{theorem}{Theorem}[section]

\theoremstyle{definition}

\theoremstyle{remark}
\newtheorem{remark}[theorem]{Remark}

\newtheorem{defn}[theorem]{Definition}

\title{An Algorithm for Computing with Brauer's Group Equivariant Neural Network Layers}

\author{%
  Edward Pearce--Crump\\
  Department of Computing\\
  Imperial College London\\
  London, SW7 2AZ, United Kingdom\\
  \texttt{ep1011@ic.ac.uk} \\
}

\begin{document}

\maketitle

\begin{abstract}
	The learnable, linear neural network layers between tensor power spaces of $\mathbb{R}^{n}$
	that are equivariant to the orthogonal group, $O(n)$, the special orthogonal group, $SO(n)$, and the symplectic group, $Sp(n)$, were characterised in \cite{pearcecrumpB}.
	We present an algorithm for multiplying a vector by any weight matrix for each of these groups, 
	using category theoretic constructions to implement the procedure.
	We achieve a significant reduction in computational cost compared with a naive implementation
	by making use of Kronecker product matrices to perform the multiplication.
	We show that our approach extends to the symmetric group, $S_n$, recovering the algorithm of \cite{godfrey} in the process.
\end{abstract}

\section{Introduction}

There has been an increased focus in deep learning to develop neural network architectures that are equivariant to a symmetry group.
When we use such a neural network, we know exactly how the output changes when a symmetry transformation is applied to the input.
These neural networks come with additional benefits: they require less training data;
the layers themselves have a high level of parameter sharing; and
there is also a reduction in the time, effort and cost that is needed to search for a neural network architecture,
since the form of the architectures is restricted by the symmetry group itself.

\cite{pearcecrumpB} recently characterised the learnable, linear neural network layers between tensor power spaces of $\mathbb{R}^{n}$ that are equivariant to the orthogonal group, $O(n)$, the special orthogonal group, $SO(n)$, and the symplectic group, $Sp(n)$. 
In particular, they found a spanning set of matrices
that are indexed by certain sets of set partition diagrams
for the learnable, linear, equivariant layer functions between such tensor power spaces in the standard basis of $\mathbb{R}^{n}$ when the group is $O(n)$ or $SO(n)$, and in the symplectic basis of $\mathbb{R}^{n}$ when the group is $Sp(n)$.
This overparameterization of the layer spaces makes it possible to learn the weights that appear in such neural networks.

The main contribution of this paper is that we present an algorithm for multiplying any input vector by any weight matrix for each of the groups in question.
In particular, we apply the category theoretic constructions introduced in \cite{pearcecrumpC}, which build a functoral correspondence between set partition diagrams and the spanning set matrices, to work with the set partition diagrams as a proxy for the matrices themselves. 
There are three key properties that we take advantage of to develop the algorithm.
The first is that the functors are \textit{full}. 
This means that, in our case, we can recover the spanning set from the corresponding set of set partition diagrams.
The second is that the set partition categories, in which the set partition diagrams form the morphisms in the category, are (strict) \textit{monoidal}. 
This means that not only is it possible to manipulate the connected components of the set partition diagrams as if they were strings, with the potential to form new set partition diagrams, but also certain set partition diagrams can be decomposed into a tensor product of smaller set partition diagrams.
We focus in particular on trying to construct \textit{planar} set partition diagrams -- diagrams where no connected components intersect each other -- as they decompose into the smallest possible set partition diagrams.
The third is that the functors themselves are \textit{monoidal}.
Critically, this means that any tensor product decomposition of diagrams is respected when viewed as matrices; that is, under the functor, we obtain a Kronecker product of matrices, where each matrix is indexed by a set partition diagram.
By using these properties, we construct an algorithm that achieves a significant reduction in computational cost compared with a naive implementation, since we use a Kronecker product of smaller sized matrices to perform the multiplication.
We also show that our approach extends to the symmetric group, $S_n$, recovering, with one key distinction, the algorithm of \cite{godfrey} in the process.

\section{Preliminaries}

\begin{figure}[t]
	\begin{center}
	\scalebox{0.37}{\begin{tikzpicture}
	\begin{pgfonlayer}{nodelayer}
		\node [style=none] (94) at (6, 1) {$1$};
		\node [style=none] (95) at (9, 1) {$2$};
		\node [style=none] (96) at (12, 1) {$3$};
		\node [style=none] (97) at (15, 1) {$4$};
		\node [style=none] (99) at (3, -5) {$6$};
		\node [style=none] (100) at (6, -5) {$7$};
		\node [style=none] (101) at (9, -5) {$8$};
		\node [style=none] (102) at (12, -5) {$9$};
		\node [style=none] (103) at (15, -5) {$10$};
		\node [style=none] (104) at (2.5, 0.5) {};
		\node [style=none] (105) at (2.5, -4.5) {};
		\node [style=none] (106) at (21.5, -4.5) {};
		\node [style=none] (107) at (21.5, 0.5) {};
		\node [style=node] (108) at (6, 0) {};
		\node [style=node] (109) at (9, 0) {};
		\node [style=node] (110) at (12, 0) {};
		\node [style=node] (111) at (15, 0) {};
		\node [style=node] (112) at (3, -4) {};
		\node [style=node] (113) at (6, -4) {};
		\node [style=node] (114) at (9, -4) {};
		\node [style=node] (115) at (12, -4) {};
		\node [style=node] (116) at (15, -4) {};
		\node [style=node] (117) at (18, -4) {};
		\node [style=none] (118) at (30, 1) {$1$};
		\node [style=none] (119) at (33, 1) {$2$};
		\node [style=none] (120) at (36, 1) {$3$};
		\node [style=none] (121) at (39, 1) {$4$};
		\node [style=none] (123) at (27, -5) {$6$};
		\node [style=none] (124) at (30, -5) {$7$};
		\node [style=none] (125) at (33, -5) {$8$};
		\node [style=none] (126) at (36, -5) {$9$};
		\node [style=none] (127) at (39, -5) {$10$};
		\node [style=node] (132) at (30, 0) {};
		\node [style=node] (133) at (33, 0) {};
		\node [style=node] (134) at (36, 0) {};
		\node [style=node] (135) at (39, 0) {};
		\node [style=node] (136) at (27, -4) {};
		\node [style=node] (137) at (30, -4) {};
		\node [style=node] (138) at (33, -4) {};
		\node [style=node] (139) at (36, -4) {};
		\node [style=node] (140) at (39, -4) {};
		\node [style=node] (141) at (42, -4) {};
		\node [style=none] (142) at (26.5, 0.5) {};
		\node [style=none] (143) at (26.5, -4.5) {};
		\node [style=none] (144) at (45.5, -4.5) {};
		\node [style=none] (145) at (45.5, 0.5) {};
		\node [style=none] (146) at (0, -2) {\Huge$\bm{b)}$};
		\node [style=none] (147) at (24, -2) {\Huge$\bm{c)}$};
		\node [style=none] (148) at (-18, 1) {$1$};
		\node [style=none] (149) at (-15, 1) {$2$};
		\node [style=none] (150) at (-12, 1) {$3$};
		\node [style=none] (151) at (-9, 1) {$4$};
		\node [style=none] (152) at (-6, 1) {$5$};
		\node [style=none] (153) at (-21, -5) {$6$};
		\node [style=none] (154) at (-18, -5) {$7$};
		\node [style=none] (155) at (-15, -5) {$8$};
		\node [style=none] (156) at (-12, -5) {$9$};
		\node [style=none] (157) at (-9, -5) {$10$};
		\node [style=none] (158) at (-21.5, 0.5) {};
		\node [style=none] (159) at (-21.5, -4.5) {};
		\node [style=none] (160) at (-2.5, -4.5) {};
		\node [style=none] (161) at (-2.5, 0.5) {};
		\node [style=node] (162) at (-18, 0) {};
		\node [style=node] (163) at (-15, 0) {};
		\node [style=node] (164) at (-12, 0) {};
		\node [style=node] (165) at (-9, 0) {};
		\node [style=node] (166) at (-21, -4) {};
		\node [style=node] (167) at (-18, -4) {};
		\node [style=node] (168) at (-15, -4) {};
		\node [style=node] (169) at (-12, -4) {};
		\node [style=node] (170) at (-9, -4) {};
		\node [style=node] (171) at (-6, -4) {};
		\node [style=none] (172) at (-24, -2) {\Huge$\bm{a)}$};
		\node [style=node] (173) at (-3, -4) {};
		\node [style=none] (174) at (-6, -5) {$11$};
		\node [style=node] (175) at (21, -4) {};
		\node [style=none] (176) at (18, -5) {$11$};
		\node [style=node] (177) at (45, -4) {};
		\node [style=none] (178) at (42, -5) {$11$};
		\node [style=node] (179) at (-6, 0) {};
		\node [style=none] (180) at (-3, -5) {$12$};
		\node [style=none] (181) at (42, 1) {$5$};
		\node [style=node] (182) at (42, 0) {};
		\node [style=none] (183) at (18, 1) {$5$};
		\node [style=node] (184) at (18, 0) {};
		\node [style=none] (185) at (21, -5) {$12$};
		\node [style=none] (186) at (45, -5) {$12$};
	\end{pgfonlayer}
	\begin{pgfonlayer}{edgelayer}
		\draw [style={dash_2}] (105.center)
			 to (106.center)
			 to (107.center)
			 to (104.center)
			 to cycle;
		\draw [style=thick] (112) to (108);
		\draw [style=thick, bend left=60, looseness=1.25] (113) to (114);
		\draw [style=thick] (115) to (109);
		\draw [style=thick, bend right=60, looseness=1.25] (110) to (111);
		\draw [style=thick, bend left=60, looseness=1.25] (116) to (117);
		\draw [style={dash_3}] (143.center)
			 to (144.center)
			 to (145.center)
			 to (142.center)
			 to cycle;
		\draw [style=thick] (137) to (133);
		\draw [style=thick, bend left=60, looseness=1.25] (139) to (140);
		\draw [style={dash_4}] (161.center)
			 to (158.center)
			 to (159.center)
			 to (160.center)
			 to cycle;
		\draw [style=thick] (168) to (164);
		\draw [style=thick] (162) to (169);
		\draw [style=thick, bend left=60, looseness=1.25] (170) to (171);
		\draw [style=thick] (170) to (165);
		\draw [style=thick] (177) to (182);
		\draw [style=thick] (175) to (184);
		\draw [style=thick] (173) to (179);
		\draw [style=thick, bend left=300, looseness=1.25] (168) to (167);
		\draw [style=thick] (166) to (163);
	\end{pgfonlayer}
\end{tikzpicture}}
		\caption{Examples of $(7,5)$--partition diagrams. b) is also a $(7,5)$--Brauer diagram, and c) is also a $12 \backslash 6$--diagram.}
	\label{partdiagrams1}
	\end{center}
\end{figure}
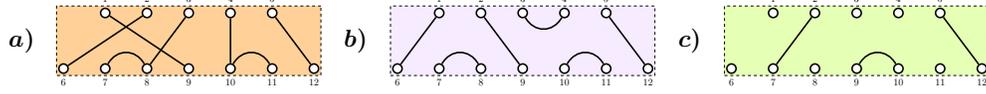

We choose our field of scalars to be $\mathbb{R}$ throughout. 
Tensor products are also taken over $\mathbb{R}$, unless otherwise stated.
Also, we let $[n]$ represent the set $\{1, \dots, n\}$. 

Recall that a representation of a group $G$ is a choice of vector space $V$ over $\mathbb{R}$ and a group homomorphism
\begin{equation} \label{grouprephom}
	\rho_V : G \rightarrow GL(V)	
\end{equation}
Furthermore, recall that a map $\phi : V \rightarrow W$ between two representations of $G$ is said to be $G$-equivariant if, for all $g \in G$ and $v \in V$, 
\begin{equation} \label{Gequivmapdefn}
	\phi(\rho_{V}(g)[v]) = \rho_{W}(g)[\phi(v)]
\end{equation}
We denote the set of all \textit{linear} $G$-equivariant maps between $V$ and $W$ by $\Hom_{G}(V,W)$. 
It can be shown that $\Hom_{G}(V,W)$ is a vector space over $\mathbb{R}$. 
See \cite{segal} for more details. 

\subsection{Tensor Power Spaces as Group Representations} \label{tenspowerspaces}

The groups $O(n)$, $Sp(n)$, and $SO(n)$
are subgroups of $GL(n)$. We use the symbol $G$ to refer to any of these groups in the following.
Recall that $\mathbb{R}^{n}$ has a standard basis that is given by $\{e_i \mid i \in [n]\}$, where $e_i$ has a $1$ in the $i^{\text{th}}$ position and is $0$ otherwise.

(Note that if $G = Sp(n)$, then $n = 2m$, and we label and order the indices by
$1, 1', \dots, m, m'$,
and call the standard basis of $\mathbb{R}^{n}$ the symplectic basis.)

There exists a (left) action of $G$ on $\mathbb{R}^{n}$ that is given by left multiplication on the standard basis, which can be extended linearly to obtain a representation $G \rightarrow GL(\mathbb{R}^n)$.

Moreover, 
since the elements 
\begin{equation} \label{tensorelementfirst}
	e_I \coloneqq e_{i_1} \otimes e_{i_2} \otimes \dots \otimes e_{i_k} 
\end{equation}
for all $I \coloneqq (i_1, i_2, \dots, i_k) \in [n]^k$ form a basis of 
$(\mathbb{R}^{n})^{\otimes k}$,
the $k$--tensor power space of $\mathbb{R}^{n}$,
there also exists a (left) action of $G$ on $(\mathbb{R}^{n})^{\otimes k}$ that is given by
\begin{equation} 
	g \cdot e_I \coloneqq ge_{i_1} \otimes ge_{i_2} \otimes \dots \otimes ge_{i_k} 
\end{equation}
Again, this action can be extended linearly to obtain a representation $\rho_k: G \rightarrow GL((\mathbb{R}^n)^{\otimes k})$.

We are interested in the space of $G$--equivariant linear maps
between any two tensor power spaces of $\mathbb{R}^{n}$, 
$\Hom_G((\mathbb{R}^{n})^{\otimes k}, (\mathbb{R}^{n})^{\otimes l})$,
since these maps are the linear layer functions in the group equivariant neural networks of interest.

\subsection{Set Partition Categories} \label{setpartdiag}

\cite{pearcecrump, pearcecrumpB} showed that, for the groups $G$ in question, 
$\Hom_G((\mathbb{R}^{n})^{\otimes k}, (\mathbb{R}^{n})^{\otimes l})$
can be constructed from certain set partitions of $[l+k]$, and in particular, from their corresponding set partition diagrams. 
\cite{pearcecrumpC} introduced a category theoretic framework around these set partition diagrams which allows us
to better understand and work with the linear layer functions of the neural networks themselves.
We assume throughout that $n \in \mathbb{N}_{\geq 0}$.

For $l, k \in \mathbb{N}_{\geq 0}$, consider the set 
$[l+k] \coloneqq \{1, \dots, l+k\}$
having $l+k$ elements. 
We can create a set partition of $[l+k]$ by partitioning it into a number of subsets.
We call the subsets of a set partition \textit{blocks}.
Let $\Pi_{l+k}$ be the set of all set partitions of $[l+k]$.
Then, for each set partition $\pi$ in $\Pi_{l+k}$, we can associate to it a diagram $d_\pi$,
called a $(k,l)$--partition diagram,
consisting of two
rows of vertices and edges between vertices such that there are
\begin{itemize}
	\item $l$ vertices on the top row, labelled left to right by $1, \dots, l$
	\item $k$ vertices on the bottom row, labelled left to right by $l+1, \dots, l+k$, and
	\item the edges between the vertices correspond to the connected components of $\pi$. 
\end{itemize}
As a result, $d_\pi$ represents the equivalence class of all diagrams with connected components equal to the blocks of $\pi$.

There are special types of $(k,l)$--partition diagrams that we are interested in, namely:
\begin{itemize}
	\item A $(k,l)$--Brauer diagram $d_\beta$ is a $(k,l)$--partition diagram where the size of every block in $\beta$ is exactly two.
	\item Given $k$ and $l$, an $(l+k)\backslash n$--diagram $d_\alpha$ is a $(k,l)$--partition diagram where exactly $n$ blocks in $\alpha$ have size one, with the rest having exactly size two. The vertices corresponding to the blocks of size one are called free vertices.
\end{itemize}
We give examples of these set partition diagrams in Figure \ref{partdiagrams1}.

From these special types of set partition diagrams, we can form a number of 
set partition categories, as follows.
\begin{defn}
	The Brauer category $\mathcal{B}(n)$ is the category
	whose objects are the non--negative integers $\mathbb{N}_{\geq 0} = \{0, 1, 2, \dots \}$,
	and, for any pair of objects $k$ and $l$, the morphism space
	$\Hom_{\mathcal{B}(n)}(k,l)$ 
	is a vector space that is defined to be the $\mathbb{R}$-linear span of the set of all $(k,l)$--Brauer diagrams.
\end{defn}

\begin{defn}
	The Brauer--Grood category $\mathcal{BG}(n)$ is the category
	whose objects are the same as those of $\mathcal{B}(n)$
	and, for any pair of objects $k$ and $l$, the morphism space
	$\Hom_{\mathcal{BG}(n)}(k,l)$ is defined to be the $\mathbb{R}$-linear span of the set of all $(k,l)$--Brauer diagrams together with the set of all $(l+k)\backslash n$--diagrams.
\end{defn}

These two categories come with a vertical composition operation on morphisms, a tensor product operation on objects and morphisms, and a unit object.
The vertical composition operation can be found in \cite[Section 2.2, Appendix A]{pearcecrumpC}.
The unit object in each category is the object $0$.
The tensor product operation on objects is given by the standard addition operation in $\mathbb{N}_{\geq 0}$.
Finally, the tensor product operation is defined on diagrams (morphisms) as follows:

\begin{itemize}
	\item If 
$d_{\beta_1}$ is a $(k,l)$--Brauer diagram
and
$d_{\beta_2}$ is a $(q,m)$--Brauer diagram,
then 
$d_{\beta_1} \otimes d_{\beta_2}$ 
is defined to be the $(k+q,l+m)$--Brauer diagram
obtained by horizontally placing
$d_{\beta_1}$
to the left of
$d_{\beta_2}$
without any overlapping of vertices.
	\item If 
$d_{\beta}$ is a $(k,l)$--Brauer diagram
and
$d_{\alpha}$ is an $(m+q) \backslash n$--diagram,
then
$d_{\beta} \otimes d_{\alpha}$ 
is defined to be the $(l+m+k+q) \backslash n$--diagram
obtained by horizontally placing
$d_{\beta}$
to the left of
$d_{\alpha}$
without any overlapping of vertices.
Similarly for 
$d_{\alpha} \otimes d_{\beta}$.
	\item See 
\cite[Appendix A]{pearcecrumpC}
for the definition of the tensor product of an
$(l+k) \backslash n$--diagram
with an
$(m+q) \backslash n$--diagram.
\end{itemize}

$\mathcal{B}(n)$ and $\mathcal{BG}(n)$ are, in fact, strict $\mathbb{R}$--linear monoidal categories -- see \cite[Section 4.1]{pearcecrumpC} for more details.
Morphisms in such categories can be represented using a diagrammatic language known as string diagrams. 
See \cite[Section 3.2]{pearcecrumpC} for more details.
This has the consequence that we can pull on and bend the connected components as if they were strings and/or move the vertices to obtain new set partition diagrams, and hence new morphisms, in the appropriate categories.

\subsection{Group Equivariant Linear Layers} 
\label{groupequivlinlayers}

For each group $G$, there is a spanning set for
$\Hom_G((\mathbb{R}^{n})^{\otimes k}, (\mathbb{R}^{n})^{\otimes l})$
that is indexed by certain set partitions of $[l+k]$ that correspond to the special types 
of $(k,l)$--partition diagrams that were introduced in Section \ref{setpartdiag}.
These spanning sets are 
expressed in the basis of matrix units for
$\Hom((\mathbb{R}^{n})^{\otimes k}, (\mathbb{R}^{n})^{\otimes l})$.
We state here what these spanning sets are, leaving their explicit definitions to the Technical Appendix.


\begin{theorem}
	[Spanning set when $G = O(n)$]
	\cite[Theorem 6.5]{pearcecrumpB}
	\label{spanningsetO(n)}

	For any $k, l \in \mathbb{N}_{\geq 0}$ and any
	$n \in \mathbb{N}_{\geq 1}$,
	the set
	\begin{equation} \label{klOnSpanningSet}
		\{E_\beta \mid d_\beta \text{ is a } (k,l) \text{--Brauer diagram} \}
	\end{equation}
	is a spanning set for
	$\Hom_{O(n)}((\mathbb{R}^{n})^{\otimes k}, (\mathbb{R}^{n})^{\otimes l})$
	in the standard basis of $\mathbb{R}^{n}$.
\end{theorem}

\begin{theorem} 
	[Spanning set when $G = Sp(n), n = 2m$]
	\cite[Theorem 6.6]{pearcecrumpB}
	\label{spanningsetSp(n)}

	For any $k, l \in \mathbb{N}_{\geq 0}$ and any
	$n \in \mathbb{N}_{\geq 2}$ such that $n = 2m$,
	the set
	\begin{equation} \label{klSpnSpanningSet}
		\{F_\beta \mid d_\beta \text{ is a } (k,l) \text{--Brauer diagram} \}
	\end{equation}
	is a spanning set for
	$\Hom_{Sp(n)}((\mathbb{R}^{n})^{\otimes k}, (\mathbb{R}^{n})^{\otimes l})$, for $n = 2m$,
	in the symplectic basis of $\mathbb{R}^{n}$.
		

\end{theorem}

\begin{theorem} 
	[Spanning set when $G = SO(n)$]
	\cite[Theorem 6.7]{pearcecrumpB}
	\label{spanningsetSO(n)}

	For any $k, l \in \mathbb{N}_{\geq 0}$ and any $n \in \mathbb{N}_{\geq 1}$,
 	the set
	\begin{equation} \label{SOn2SpanningSet}
		\{E_\beta
		\mid \beta \text{ is a } (k,l) \text{--Brauer diagram} \} 
		\cup
		\{H_\alpha
		\mid \alpha \text{ is a } (k+l) \backslash n \text{--diagram} \} 
	\end{equation}
	is a spanning set for
	$\Hom_{SO(n)}((\mathbb{R}^{n})^{\otimes k}, (\mathbb{R}^{n})^{\otimes l})$
	in the standard basis of $\mathbb{R}^{n}$.
\end{theorem}

For each group $G$ in question, we can also define the following category.
\begin{defn} \label{catgroupreps}
	The category $\mathcal{C}(G)$ 
	consists of 
	objects that are the $k$-order tensor power spaces of $\mathbb{R}^{n}$, as representations of $G$, and morphism spaces between any two objects that are the vector spaces
$\Hom_{G}((\mathbb{R}^{n})^{\otimes k}, (\mathbb{R}^{n})^{\otimes l})$.

The vertical composition of morphisms is given by the usual composition of linear maps, the tensor product is given by the usual tensor product of linear maps, 
and the unit object is the one-dimensional trivial representation of $G$.
\end{defn}
$\mathcal{C}(G)$
is a strict, $\mathbb{R}$--linear monoidal category: see \cite[Appendix D]{pearcecrumpC} for more details.

\subsection{Full, Strict $\mathbb{R}$--Linear Monoidal Functors}

\cite[Section 4.2, Appendix D]{pearcecrumpC} showed that we have a number of 
\textit{full, strict $\mathbb{R}$--linear monoidal}
functors between the set partition categories and the category $\mathcal{C}(G)$ for the appropriate group $G$.
We reproduce the results below.

\begin{theorem} \label{brauerO(n)functor}
	There exists a full, strict $\mathbb{R}$--linear monoidal functor
	\begin{equation}
		\Phi : \mathcal{B}(n) \rightarrow \mathcal{C}(O(n))
	\end{equation}
	that is defined on the objects of $\mathcal{B}(n)$ by 
	$\Phi(k) \coloneqq ((\mathbb{R}^{n})^{\otimes k}, \rho_k)$
	and, for any objects $k,l$ of $\mathcal{B}(n)$, the map
	\begin{equation}	
		\Hom_{\mathcal{B}(n)}(k,l) 
		\rightarrow 
		\Hom_{\mathcal{C}(O(n))}(\Phi(k),\Phi(l))
	\end{equation}
	is given by
	\begin{equation}
		d_\beta \mapsto E_\beta
	\end{equation}
	for all $(k,l)$--Brauer diagrams $d_\beta$,
	where $E_\beta$ is given in Theorem \ref{spanningsetO(n)}.
\end{theorem}

\begin{theorem} \label{brauerSp(n)functor}
	There exists a full, strict $\mathbb{R}$--linear monoidal functor
	\begin{equation}
		X : \mathcal{B}(n) \rightarrow \mathcal{C}(Sp(n))
	\end{equation}
	that is defined on the objects of $\mathcal{B}(n)$ by 
	$X(k) \coloneqq ((\mathbb{R}^{n})^{\otimes k}, \rho_k)$
	and, for any objects $k,l$ of $\mathcal{B}(n)$, the map
	\begin{equation}	
		\Hom_{\mathcal{B}(n)}(k,l) 
		\rightarrow 
		\Hom_{\mathcal{C}(Sp(n))}(\Phi(k),\Phi(l))
	\end{equation}
	is given by
	\begin{equation}
		d_\beta \mapsto F_\beta
	\end{equation}
	for all $(k,l)$--Brauer diagrams $d_\beta$,
	where $F_\beta$ is given in Theorem \ref{spanningsetSp(n)}.
\end{theorem}

\begin{theorem} \label{brauerSO(n)functor}
	There exists a full, strict $\mathbb{R}$--linear monoidal functor
	\begin{equation}
		\Psi : \mathcal{BG}(n) \rightarrow \mathcal{C}(SO(n))
	\end{equation}
	that is defined on the objects of $\mathcal{BG}(n)$ by 
	$\Psi(k) \coloneqq ((\mathbb{R}^{n})^{\otimes k}, \rho_k)$
	and, for any objects $k,l$ of $\mathcal{B}(n)$, the map
	\begin{equation}	
		\Hom_{\mathcal{BG}(n)}(k,l) 
		\rightarrow 
		\Hom_{\mathcal{C}(SO(n))}(\Phi(k),\Phi(l))
	\end{equation}
	is given by
	\begin{equation}
		d_\beta \mapsto E_\beta
	\end{equation}
	for all $(k,l)$--Brauer diagrams $d_\beta$,
	where $E_\beta$ is given in Theorem \ref{spanningsetO(n)}, and
	\begin{equation}
		d_\alpha \mapsto H_\beta
	\end{equation}
	for all $(l+k) \backslash n$--diagrams $d_\alpha$,
	where $H_\alpha$ is given in Theorem \ref{spanningsetSO(n)}.
\end{theorem}

The key implications of these results going forward are as follows:
\begin{enumerate}
	\item To understand and work with any matrix in 
	$\Hom_{G}((\mathbb{R}^{n})^{\otimes k}, (\mathbb{R}^{n})^{\otimes l})$,
it is enough to work with the subset of $(k,l)$--partition diagrams that correspond to $G$.
		This is because 
		we can express any matrix in terms of the set of spanning set elements for
	$\Hom_{G}((\mathbb{R}^{n})^{\otimes k}, (\mathbb{R}^{n})^{\otimes l})$,
		given in Theorems \ref{spanningsetO(n)} -- \ref{spanningsetSO(n)}, and these correspond
		bijectively with the subset of $(k,l)$--partition diagrams that corresponds to $G$. 
		We can recover the matrix itself by 
		applying the appropriate functor to the set partition diagrams because the functors are full.
	\item We can manipulate the connected components and vertices of 
		$(k,l)$--partition diagrams like strings
		to obtain new set partition diagrams
		because the set partition categories are strict monoidal. 
	Point 1. immediately implies that we will obtain new $G$--equivariant matrices between tensor power spaces of $\mathbb{R}^{n}$ from the resulting set partition diagrams.

	\item If a $(k,l)$--partition diagram can be decomposed as a tensor product of smaller set partition diagrams, then the corresponding matrix can also be decomposed as a tensor product of smaller sized matrices, each of which is $G$--equivariant. This is because the functors are monoidal.
		It is this property that makes these specific functors so valuable in what follows, as without it, 
		using the set partition diagrams to factor the matrices will not be possible.
	\item In particular, $(k,l)$--partition diagrams that are \textit{planar}
-- that is, none of the connected components in the diagram intersect each other -- 
		can be decomposed as a tensor product of smaller set partition diagrams.
\end{enumerate}



\section{Multiplication Algorithm}

We can use the summary points given above
to construct
an algorithm for multiplying any vector
$v \in (\mathbb{R}^{n})^{\otimes k}$
by any matrix in
$\Hom_{G}((\mathbb{R}^{n})^{\otimes k}, (\mathbb{R}^{n})^{\otimes l})$,
expressed in the standard basis of $\mathbb{R}^{n}$,
for each of the groups $G$ in question.

Since we have 
a spanning set of
$\Hom_{G}((\mathbb{R}^{n})^{\otimes k}, (\mathbb{R}^{n})^{\otimes l})$ for each group $G$,
it is enough to describe an algorithm for how to multiply $v$ by a spanning set
element, since we can extend the result by linearity.
The linearity is particularly nice as it allows for the computation 
with a generic matrix in 
$\Hom_{G}((\mathbb{R}^{n})^{\otimes k}, (\mathbb{R}^{n})^{\otimes l})$
to be executed in parallel
on each of the spanning set
elements that appear in its expression.

Algorithm \ref{alg1} outlines a procedure \textsc{MatrixMult} for 
how to multiply $v$ by
a spanning set
element in 
$\Hom_{G}((\mathbb{R}^{n})^{\otimes k}, (\mathbb{R}^{n})^{\otimes l})$.
We assume that we have the set partition diagram that is associated with the spanning set
element. 
(Note that in the description of the algorithm, we have used $d_\pi$ to represent a generic $(k,l)$--partition diagram; however, the type of set partition diagrams that we can use as input depends entirely upon the group $G$.)

\begin{algorithm}
	\caption{$G$-Equivariant Linear Layer Mapping}
	\label{alg1}
  \begin{algorithmic}
	  \Procedure{MatrixMult}{$G, d_\pi, v$}
	  \State \textbf{Inputs:} $G$ is a group; $d_\pi$ is an appropriate $(k,l)$--partition diagram for $G$; $v \in (\mathbb{R}^{n})^{\otimes k}$.
      \State $\sigma_k, d_\pi, \sigma_l \gets \textsc{Factor}(G, d_\pi)$
	  \Comment{The output diagram $d_\pi$ is planar.}
      \State $v \gets \textsc{Permute}(v, \sigma_k)$
	  \Comment{$v \in (\mathbb{R}^{n})^{\otimes k}$, $\sigma_k \in S_k$}
      \State $w \gets \textsc{PlanarMult}(G, d_\pi, v)$
      \State $w \gets \textsc{Permute}(w, \sigma_l)$
	  \Comment{$w \in (\mathbb{R}^{n})^{\otimes l}$, $\sigma_l \in S_l$}
      \State \textbf{Output:} $w$
    \EndProcedure
  \end{algorithmic}
\end{algorithm}

We describe each of the procedures that appear in Algorithm \ref{alg1} in more detail below.


\textsc{Factor} is a procedure that takes as input a set partition diagram corresponding to $G$ and uses the string-like property of these diagrams to output three diagrams whose composition is equivalent to the original input diagram.
The first is a diagram that corresponds to a permutation $\sigma$, where if $i$ is a vertex in the top row, then the vertex in the bottom row that is connected to it
(using the same labelling of the vertices as the top row) is $\sigma(i)$;
the second is another set partition diagram (of the same type as the input) that is planar,
and
the third is a diagram that corresponds to another permutation,
to be interpreted in the same way as the first.

\textsc{Permute} takes as input a vector
$w \in (\mathbb{R}^{n})^{\otimes m}$,
for some $n, m$, 
that is expressed in the standard basis of $\mathbb{R}^{n}$,
and a permutation $\sigma$ in $S_m$,
and outputs another vector in 
$(\mathbb{R}^{n})^{\otimes m}$ where only the indices of the basis vectors (and not the indices of the coefficients of $w$) have been permuted according to $\sigma$.
Expressed differently, \textsc{Permute} performs the following operation, which is extended linearly:
\begin{equation} \label{permuteop}
	\sigma \cdot w_Ie_I \coloneqq w_I(e_{i_{\sigma(1)}} \otimes e_{i_{\sigma(2)}} \otimes \dots \otimes e_{i_{\sigma(m)}})
\end{equation}

\textsc{PlanarMult} takes as input a planar set partition diagram 
and a vector, and performs a fast matrix multiplication on this vector.
Since the set partition diagram is planar, we first use the monoidal property of the category in which it is a morphism to decompose it as a tensor product of smaller set partition diagrams.
Next, we apply the appropriate monoidal functor to express the matrix that the planar set partition diagram corresponds to as a Kronecker product of smaller matrices.
Finally, we perform matrix multiplication by applying these smaller matrices to the input vector from "right--to--left, diagram--by--diagram" -- to be described in more detail for each group below -- returning another vector as output.


The four step procedure given in Algorithm \ref{alg1} to perform the matrix multiplication is quicker than just performing the multiplication with the matrix as given, since we are taking advantage of the Kronecker product decomposition of a matrix in \textsc{PlanarMult} to speed up the computation.
We analyse the performance of the algorithm for each group $G$ in the Technical Appendix. 

We note that the implementation of the \textsc{Factor} and \textsc{PlanarMult} procedures vary according to the group $G$ and the type of set partition diagrams that correspond to $G$,
although they share many commonalities.
We describe the implementation of these procedures for each of the groups below.

\subsection{Orthogonal Group $O(n)$}

In this case, 
we wish to perform matrix multiplication between 
$E_\beta \in \Hom_{O(n)}((\mathbb{R}^{n})^{\otimes k}, (\mathbb{R}^{n})^{\otimes l})$
and
$v \in (\mathbb{R}^{n})^{\otimes k}$.

\begin{figure}
  \centering
	\scalebox{0.33}{\input{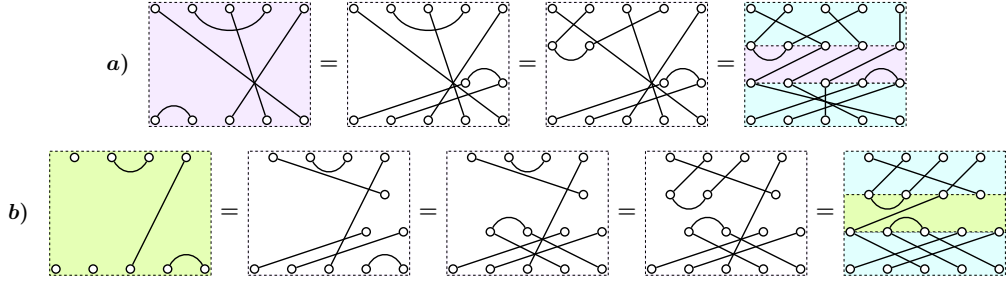}}
	\caption{a) We use the string-like aspect of $(k,l)$--Brauer diagrams to \textsc{Factor} them as a composition of a permutation in $S_k$, a \textit{planar} $(k,l)$--Brauer diagram, and a permutation in $S_l$. b) We perform the same procedure but on $(l+k) \backslash n$--diagrams.}
	\label{symmfactoringOn}
\end{figure}

\textsc{Factor}: The input is $d_\beta$,
the $(k,l)$--Brauer diagram corresponding to $E_\beta$.
We drag and bend the strings representing the connected components of $d_\beta$ to obtain a factoring of $d_\beta$ into three diagrams whose composition is equivalent to $d_\beta$:
a $(k,k)$--Brauer diagram that represents a permutation $\sigma_k$ in the symmetric group $S_k$; another $(k,l)$--Brauer diagram that is planar; and a $(l,l)$--Brauer diagram that represents a permutation $\sigma_l$ in the symmetric group $S_l$.
To obtain the desired planar $(k,l)$--Brauer diagram, 
we drag and bend the strings in any way such that 
\begin{itemize}
	\item the pairs that are solely in the bottom row of $d_\beta$ are pulled up to be next to each other
		in the far right hand side of the bottom row of the planar $(k,l)$--Brauer diagram,
	\item the pairs that are solely in the top row of $d_\beta$ are pulled down to be next to each other
		in the far left hand side of the top row of the planar $(k,l)$--Brauer diagram, and
	\item the pairs between vertices in different rows of $d_\beta$ are bent to be in between the other vertices of the planar $(k,l)$--Brauer diagram such that no two pairings in the planar diagram intersect each other.
\end{itemize}
We give an example of this procedure in Figure \ref{symmfactoringOn}a).

\begin{figure}
	\centering
	\scalebox{0.33}{\begin{tikzpicture}
	\begin{pgfonlayer}{nodelayer}
		\node [style=node] (0) at (0, -3) {};
		\node [style=node] (1) at (12, -3) {};
		\node [style=node] (2) at (9, -3) {};
		\node [style=node] (3) at (5.75, -3) {};
		\node [style=node] (4) at (6, 0) {};
		\node [style=node] (5) at (3, 0) {};
		\node [style=node] (6) at (0, 0) {};
		\node [style=node] (7) at (3, -3) {};
		\node [style=none] (10) at (-0.5, 0.5) {};
		\node [style=none] (11) at (-0.5, -3.5) {};
		\node [style=none] (12) at (12.5, -3.5) {};
		\node [style=none] (13) at (12.5, 0.5) {};
		\node [style=none] (14) at (14, -1.5) {\Huge$\bm{=}$};
		\node [style=node] (15) at (16, 0) {};
		\node [style=node] (16) at (34, -3) {};
		\node [style=node] (17) at (31, -3) {};
		\node [style=node] (18) at (28, -3) {};
		\node [style=node] (19) at (25, 0) {};
		\node [style=node] (20) at (22, 0) {};
		\node [style=node] (21) at (28, 0) {};
		\node [style=node] (22) at (25, -3) {};
		\node [style=none] (23) at (15.5, -3.5) {};
		\node [style=none] (24) at (15.5, 0.5) {};
		\node [style=none] (25) at (19.5, 0.5) {};
		\node [style=none] (26) at (19.5, -3.5) {};
		\node [style=none] (27) at (34.5, -3.5) {};
		\node [style=none] (28) at (30.5, -3.5) {};
		\node [style=none] (29) at (34.5, 0.5) {};
		\node [style=none] (30) at (30.5, 0.5) {};
		\node [style=none] (35) at (28.5, -3.5) {};
		\node [style=none] (36) at (21.5, -3.5) {};
		\node [style=none] (37) at (28.5, 0.5) {};
		\node [style=none] (38) at (21.5, 0.5) {};
		\node [style=node] (39) at (9, 0) {};
		\node [style=node] (40) at (22, -3) {};
		\node [style=node] (41) at (12, 0) {};
		\node [style=node] (42) at (19, 0) {};
		\node [style=node] (43) at (41, -3) {};
		\node [style=node] (44) at (47, -3) {};
		\node [style=node] (45) at (44, -3) {};
		\node [style=node] (46) at (52.75, -3) {};
		\node [style=node] (47) at (48.5, 0) {};
		\node [style=node] (48) at (45.5, 0) {};
		\node [style=node] (49) at (42.5, 0) {};
		\node [style=node] (50) at (50, -3) {};
		\node [style=none] (51) at (40.5, 0.5) {};
		\node [style=none] (52) at (40.5, -3.5) {};
		\node [style=none] (53) at (53.5, -3.5) {};
		\node [style=none] (54) at (53.5, 0.5) {};
		\node [style=none] (55) at (55, -1.5) {\Huge$\bm{=}$};
		\node [style=node] (56) at (51.5, 0) {};
		\node [style=node] (57) at (57, 0) {};
		\node [style=node] (58) at (69, -3) {};
		\node [style=node] (59) at (66, -3) {};
		\node [style=node] (60) at (75, -3) {};
		\node [style=node] (61) at (73.5, 0) {};
		\node [style=node] (62) at (63, 0) {};
		\node [style=node] (63) at (72, -3) {};
		\node [style=none] (64) at (56.5, -3.5) {};
		\node [style=none] (65) at (56.5, 0.5) {};
		\node [style=none] (66) at (60.5, 0.5) {};
		\node [style=none] (67) at (60.5, -3.5) {};
		\node [style=none] (68) at (69.5, -3.5) {};
		\node [style=none] (69) at (65.5, -3.5) {};
		\node [style=none] (70) at (69.5, 0.5) {};
		\node [style=none] (71) at (65.5, 0.5) {};
		\node [style=none] (72) at (75.5, -3.5) {};
		\node [style=none] (73) at (71.5, -3.5) {};
		\node [style=none] (74) at (75.5, 0.5) {};
		\node [style=none] (75) at (71.5, 0.5) {};
		\node [style=node] (76) at (63, -3) {};
		\node [style=node] (77) at (60, 0) {};
		\node [style=none] (78) at (62.5, 0.5) {};
		\node [style=none] (79) at (63.5, 0.5) {};
		\node [style=none] (80) at (62.5, -3.5) {};
		\node [style=none] (81) at (63.5, -3.5) {};
		\node [style=none] (82) at (38, -1.5) {\Huge$\bm{b)}$};
		\node [style=none] (83) at (-3, -1.5) {\Huge$\bm{a)}$};
	\end{pgfonlayer}
	\begin{pgfonlayer}{edgelayer}
		\draw [style={dash_2}] (13.center)
			 to (10.center)
			 to (11.center)
			 to (12.center)
			 to cycle;
		\draw [style=thick] (0) to (4);
		\draw [style=thick] (7) to (39);
		\draw [style=thick] (3) to (41);
		\draw [style=thick, bend left=60, looseness=1.25] (2) to (1);
		\draw [style=thick, bend left=60, looseness=1.25] (5) to (6);
		\draw [style={dash_2}] (24.center)
			 to (23.center)
			 to (26.center)
			 to (25.center)
			 to cycle;
		\draw [style=thick, bend right=60, looseness=1.25] (15) to (42);
		\draw [style={dash_2}] (35.center)
			 to (37.center)
			 to (38.center)
			 to (36.center)
			 to cycle;
		\draw [style={dash_2}] (27.center)
			 to (29.center)
			 to (30.center)
			 to (28.center)
			 to cycle;
		\draw [style=thick, bend left=60, looseness=1.25] (17) to (16);
		\draw [style=thick] (40) to (20);
		\draw [style=thick] (22) to (19);
		\draw [style=thick] (18) to (21);
		\draw [style={dash_3}] (54.center)
			 to (53.center)
			 to (52.center)
			 to (51.center)
			 to cycle;
		\draw [style=thick] (43) to (47);
		\draw [style=thick, bend right=60, looseness=1.25] (49) to (48);
		\draw [style=thick, bend left=60, looseness=1.25] (45) to (44);
		\draw [style={dash_3}] (73.center)
			 to (75.center)
			 to (74.center)
			 to (72.center)
			 to cycle;
		\draw [style={dash_2}] (66.center)
			 to (65.center)
			 to (64.center)
			 to (67.center)
			 to cycle;
		\draw [style=thick, bend right=60, looseness=1.25] (57) to (77);
		\draw [style={dash_2}] (69.center)
			 to (71.center)
			 to (70.center)
			 to (68.center)
			 to cycle;
		\draw [style=thick, bend left=60, looseness=1.25] (59) to (58);
		\draw [style={dash_2}] (78.center)
			 to (80.center)
			 to (81.center)
			 to (79.center)
			 to cycle;
		\draw [style=thick] (76) to (62);
	\end{pgfonlayer}
\end{tikzpicture}}
	\caption{
		a) The tensor product decomposition of the planar $(5,5)$--Brauer diagram appearing in Figure \ref{symmfactoringOn}a).
		b) The tensor product decomposition of the planar $(4+5) \backslash 3$--diagram appearing in Figure \ref{symmfactoringOn}b).}
	\label{tensorproddecompOn}
\end{figure}
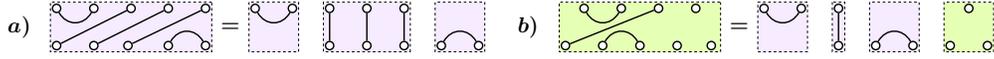

\textsc{PlanarMult}: First, we take the planar $(k,l)$--Brauer diagram that comes from \textsc{Factor} and express it as a tensor product of three types of Brauer diagrams.
The right-most type is itself a tensor product of Brauer diagrams, where each diagram has only two connected vertices in the bottom row.
The middle type is 
a Brauer diagram that consists of all of the pairs in the planar $(k,l)$--Brauer diagram between vertices in different rows.
The left-most type is a tensor product of Brauer diagrams having only two connected vertices in the top row. 
Figure \ref{tensorproddecompOn}a) shows
an example of the tensor product decomposition for the planar $(5,5)$--Brauer diagram that appears in Figure \ref{symmfactoringOn}a).

\begin{figure}
  		\centering
		\scalebox{0.33}{\input{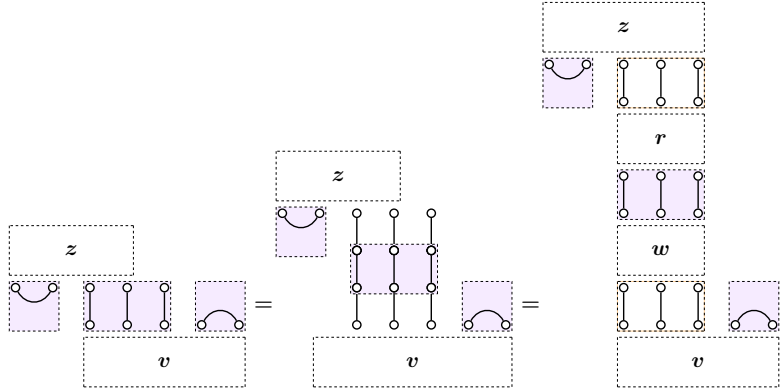}}
		\caption{We show how matrix multiplication is implemented in \textsc{PlanarMult} for $O(n), Sp(n)$ and $SO(n)$ using the tensor product decomposition of the planar 
		$(5,5)$--Brauer diagram
		given in Figure \ref{tensorproddecompOn}a)
		as an example.
		Effectively, we perform the matrix multiplication by applying the matrices "right--to--left, diagram--by--diagram".
		In reality, we perform the matrix multiplication as follows:
		first, we deform the entire tensor product decomposition diagram by pulling each individual diagram up one level higher than the previous one, going from right--to-left, and then we apply the functor that corresponds to the group at each level. 
		Finally, we perform matrix multiplication at each level to obtain the final output vector.}
		\label{planarmultOn}
	\end{figure}

The resulting tensor product decomposition corresponds to a Kronecker product of smaller matrices under the functor $\Phi$, defined in Theorem \ref{brauerO(n)functor},
by the monoidal property of $\Phi$.
In order to perform the matrix multiplication, 
we would like to apply each smaller matrix
to the input vector 
from right--to--left, diagram--by--diagram.
To do this, we first deform the entire tensor product decomposition of Brauer diagrams by pulling each individual diagram up one level higher than the previous one, going from right--to--left,
and then apply the functor $\Phi$ at each level. 
The newly inserted strings correspond to an identity matrix, hence only the matrices corresponding to the original tensor product decomposition act on the input vector at each stage, as desired!
We give an example in Figure \ref{planarmultOn} of how the computation takes place at each stage for the tensor product decomposition given in Figure \ref{tensorproddecompOn}a), using its equivalent diagram form. 
We provide full details of how this procedure is implemented in the Technical Appendix.

\begin{remark} \label{remarkfactimpl}
	It is important to highlight that the implementation of \textsc{Factor} is very important to the overall performance of the entire algorithm.
	Specifically, we want to obtain a particular planar Brauer diagram, hence the use of the word \textit{desired} above.
	Firstly, we want the middle Brauer diagram to be planar in order to take advantage of the fact that it can be decomposed as a tensor product of smaller Brauer diagrams, as this corresponds to a Kronecker product of matrices under the functor $\Phi$.
	However, when performing matrix multiplication with a Kronecker product of matrices, these matrices will perform different operations, and so, if we can choose their order, we should do so in order to execute them in the most efficient way possible.
	It is clear that ordering the matrices is equivalent to ordering the smaller diagrams, which is equivalent to obtaining a specific planar Brauer diagram!

	Under the functor $\Phi$,
	the right-most type of Brauer diagrams that are used in \textsc{PlanarMult} 
	corresponds to tensor contraction (indexing and summation) operations.
	The middle type corresponds to index transfer operations, which, in this case, act as the identity transformation -- hence no such operations are executed.
	Finally, the left-most type corresponds to indexing operations that perform copies.
	In particular, it is best to perform tensor contraction operations first before performing copying operations as we are reducing the number of elements that need to be copied; this is why we want \textsc{Factor} to return the particular planar Brauer diagram that it does.
	We analyse the performance of these operations further in the Technical Appendix.
\end{remark}

\subsection{Symplectic Group $Sp(n)$}

In this case, 
we wish to perform matrix multiplication between 
$F_\beta \in \Hom_{Sp(n)}((\mathbb{R}^{n})^{\otimes k}, (\mathbb{R}^{n})^{\otimes l})$ and
$v \in (\mathbb{R}^{n})^{\otimes k}$.

The implementation of the \textsc{Factor} procedure is the same as for the orthogonal group.
The implementation of the \textsc{PlanarMult} procedure is also the same as for the orthogonal group, except we apply the functor $X$, defined in Theorem \ref{brauerSp(n)functor}, instead of the functor $\Phi$, to perform the matrix multiplication.
Note that the three types of diagrams correspond to operations of the same nature as for the orthogonal group;
however, the horizontal pairs correspond to different matrices due to the change in functor.
We provide full details of how this procedure is implemented in the Technical Appendix.

\subsection{Special Orthogonal Group $SO(n)$}



We wish to 
perform matrix multiplication between
either
$E_\beta$ or $H_\alpha
\in \Hom_{SO(n)}((\mathbb{R}^{n})^{\otimes k}, (\mathbb{R}^{n})^{\otimes l})$
and
$v \in (\mathbb{R}^{n})^{\otimes k}$,
where $d_\beta$ is a $(k,l)$--Brauer diagram,
and $d_\alpha$ is a $(l+k) \backslash n$--diagram.
The $E_\beta$ case is the same as for the orthogonal group by Theorem \ref{spanningsetSO(n)}.
We consider the $H_\alpha$ case below.

\textsc{Factor}:
The input is $d_\alpha$.
Again, we drag and bend the strings representing the connected components of $d_\alpha$ to obtain a factoring of $d_\alpha$ into the same three diagrams, except this time the middle diagram will be a planar $(l+k) \backslash n$--diagram.
To obtain the desired planar 
$(l+k) \backslash n$--diagram
we want to drag and bend the strings in any way such that 
\begin{itemize}
	\item the free vertices in the top row of $d_\alpha$ are pulled down to the far right of the top row of the planar $(l+k) \backslash n$--diagram, maintaining their order,
	\item the free vertices in the bottom row of $d_\alpha$ are pulled up to the far right of the bottom row of the planar $(l+k) \backslash n$--diagram, maintaining their order,
	\item the pairs in the bottom row of $d_\alpha$ are pulled up to be next to each other in the right hand side of the bottom row of the planar $(l+k) \backslash n$--diagram, but next to and to the left of the free vertices in the bottom row of the planar $(l+k) \backslash n$--diagram, 
	\item the pairs in the top row of $d_\alpha$ are pulled down to be next to each other in the far left hand side of the top row of the planar $(l+k) \backslash n$--diagram,
	\item the pairs connecting vertices in different rows of $d_\alpha$ are ordered in the planar $(l+k) \backslash n$--diagram in between the other vertices such that no two pairings in the planar diagram intersect each other.
\end{itemize}
We give an example of this procedure in Figure \ref{symmfactoringOn}b).

\textsc{PlanarMult}:
The implementation is slightly different. 
Again, we take the planar
$(l+k) \backslash n$--diagram
that comes from \textsc{Factor}, but now express it as a tensor product of \textit{four} types of set partition diagrams. 
The right-most is a diagram consisting of all the free vertices. 
This corresponds to the evaluation operation $\chi$ that is given in 
\cite[Theorem 6.7]{pearcecrumpB}.
The other types of diagrams (and their corresponding operations) are exactly the same as for the orthogonal group.
Figure \ref{tensorproddecompOn}b) shows an example of this tensor product decomposition.

The matrix multiplication step is very similar to the orthogonal group, in that to obtain the matrices we perform the same 
deformation of the tensor product decomposition of diagrams 
before applying the functor $\Psi$, defined in Theorem \ref{brauerSO(n)functor}, at each level.
Note, in particular, that we need to attach identity strings to the free vertices appearing in the top row.
We provide full details of how this procedure is implemented in the Technical Appendix.

\begin{remark}
	Similar to Remark \ref{remarkfactimpl}, we want \textsc{Factor} to return a specific planar $(l+k) \backslash n$--diagram in order to obtain the most efficient matrix multiplication possible. 
	In this case, we want to pull the free vertices over to the far right hand side as such a diagram corresponds to an operation that zeroes out most terms in the input vector; in fact, by the definition of $\chi$, the number of terms will decrease from $n^{k}$ to $n!$. 
	As a consequence of the form of a planar $(l+k) \backslash n$--diagram,
	the rest of the order is determined by the order for the orthogonal group case.
\end{remark}

\begin{remark}
	The methods that we have used to construct the algorithm presented in this paper can be extended to the case where the group $G$ is the symmetric group $S_n$.
	In this case, we recover, in effect, the algorithm given in \cite[Appendix C]{godfrey}; however, we use an entirely different approach -- involving monoidal categories -- to obtain it.
	We provide full details of its implementation, as well as a discussion on some key differences between the two versions, in the Technical Appendix. 
\end{remark}	

We also give examples of how to perform \textsc{MatrixMult} for each of the groups in question in the Technical Appendix. 

\section{Related Work}

The motivation for our algorithm comes from the literature on permutation equivariant neural networks.
\cite{maron2018} were the first to classify the linear layer functions in such networks.
\cite{pearcecrump} then established a connection between 
these layer functions
and the partition algebra using Schur–Weyl duality. 
\cite{pan22} investigated the operations that are needed to perform matrix multiplication with these layers.
\cite{godfrey} implemented an algorithm to perform the matrix multiplication itself,
and found that
using a basis known as the diagram basis, first constructed by \cite{Jones} in the case $k = l$, 
is particularly beneficial for these computations.

\cite{pearcecrumpB} characterised the learnable, linear neural network layers between tensor power spaces of $\mathbb{R}^{n}$ that are equivariant to the orthogonal group, $O(n)$, the special orthogonal group, $SO(n)$, and the symplectic group, $Sp(n)$. 
The Brauer algebra, which appears in this characterisation,
was first developed by \cite{Brauer}.
\cite{brown1,brown2} showed that the Brauer algebra 
is semisimple if and only if $n \geq k-1$.
\cite{grood} investigated the representation theory of the Brauer--Grood algebra.
The Brauer category first appeared in \cite{LehrerZhang}, and it is also discussed in \cite{Hu2019}.
\cite{LehrerZhang2} investigated the theory behind what we have termed the Brauer–Grood category. This category also appears in \cite{comes}.

\section{Conclusion}

In this paper, we have introduced an algorithm for multiplying a vector by any weight matrix that appears in a group equivariant neural network where the layers are tensor power spaces of $\mathbb{R}^{n}$, for the orthogonal, special orthogonal, and symplectic groups.
Our implementation takes advantage of the properties of monoidal categories and monoidal functors, and results in a significant reduction in computational cost compared with a naive implementation. 
Ultimately, this algorithm reduces the time that is needed to train and run such neural networks.



\section{Acknowledgments}
The author would like to thank his PhD supervisor Professor William J. Knottenbelt for being generous with his time throughout the author's period of research prior to the publication of this paper.

This work was funded by the Doctoral Scholarship for Applied Research which was awarded to the author under Imperial College London's Department of Computing Applied Research scheme.
This work will form part of the author's PhD thesis at Imperial College London.
 


\nocite{*}
\bibliography{references}
\bibliographystyle{icml2023}


\newpage

\begin{appendix}

	\section{Group Equivariant Linear Layer Definitions}

	In Section \ref{groupequivlinlayers},
	we said that there exists a spanning set of
	$\Hom_{G}((\mathbb{R}^{n})^{\otimes k}, (\mathbb{R}^{n})^{\otimes l})$ for each of the groups in question, expressed in the basis of matrix units of 
$\Hom((\mathbb{R}^{n})^{\otimes k}, (\mathbb{R}^{n})^{\otimes l})$, without stating what the matrices actually are
in the main text. 
We state what these matrices are below.
	These definitions can also be found in \cite[Theorems 6.5, 6.6, 6.7]{pearcecrumpB} for $O(n), Sp(n)$ and $SO(n)$, respectively.

Recall that, for any $k,l \in \mathbb{N}_{\geq 0}$, 
as a result of picking the standard/symplectic basis for $\mathbb{R}^{n}$,
the vector space
$\Hom((\mathbb{R}^{n})^{\otimes k}, (\mathbb{R}^{n})^{\otimes l})$
has a standard basis of matrix units
\begin{equation} \label{standardbasisunits}
	\{E_{I,J}\}_{I \in [n]^l, J \in [n]^k}
\end{equation}
where $I$ is a tuple $(i_1, i_2, \dots, i_l) \in [n]^l$, 
$J$ is a tuple $(j_1, j_2, \dots, j_k) \in [n]^k$
and $E_{I,J}$ has a $1$ in the $(I,J)$ position and is $0$ elsewhere.
If one or both of $k$, $l$ is equal to $0$, then we replace the tuple that indexes the matrix by a $1$.
For example, when $k = 0$ and $l \in \mathbb{N}_{\geq 1}$, (\ref{standardbasisunits}) becomes $\{E_{I,1}\}_{I \in [n]^l}$.

	\begin{defn}[$E_\beta$ given in Theorem \ref{spanningsetO(n)}] \label{Sndiagdefn}

		Suppose that $d_\beta$ is a $(k,l)$--Brauer diagram.
		Then $E_\beta$ is defined as follows.

		Associate the indices $i_1, i_2, \dots, i_l$ with the vertices in the top row of $d_\pi$, and $j_1, j_2, \dots, j_k$ with the vertices in the bottom row of $d_\pi$.
	Then, if $S_\beta((I,J))$ is defined to be the set
	\begin{equation} \label{Snindexingset}
		\{(I,J) \in [n]^{l+k} \mid \text{if } x,y \text{ are in the same pair of } \beta, \text{then } i_x = i_y \}
	\end{equation}
	(where we have momentarily replaced the elements of $J$ by $i_{l+m} \coloneqq j_m$ for all $m \in [k]$),
	 we have that
	\begin{equation} \label{mappeddiagbasisOn}
		E_\beta
		\coloneqq
		\sum_{I \in [n]^l, J \in [n]^k}
		\delta_{\beta, (I,J)}
		E_{I,J}
	\end{equation}
	where
	\begin{equation}
		\delta_{\beta, (I,J)}
		\coloneqq
		\begin{cases}
			1 & \text{if } (I,J) \in S_\beta((I,J)) \\
			0 & \text{otherwise}
		\end{cases}
	\end{equation}
	\end{defn}

	\begin{defn}[$F_\beta$ given in Theorem \ref{spanningsetSp(n)}]
		Suppose that $d_\beta$ is a $(k,l)$--Brauer diagram.
		Then $F_\beta$ is defined as follows.
	
		Associate the indices $i_1, i_2, \dots, i_l$ with the vertices in the top row of $d_\beta$, and $j_1, j_2, \dots, j_k$ with the vertices in the bottom row of $d_\beta$.
	Then, we have that
	\begin{equation} \label{matrixSp(n)}
		F_\beta 
		\coloneqq
		\sum_{I, J} 
		\gamma_{r_1, u_1}
		\gamma_{r_2, u_2}
		\dots
		\gamma_{r_{\frac{l+k}{2}}, u_{\frac{l+k}{2}}}
		E_{I,J}
	\end{equation}
	where the indices $i_p, j_p$ range over $1, 1', \dots, m, m'$,
	where $r_1, u_1, \dots, r_{\frac{l+k}{2}}, u_{\frac{l+k}{2}}$ is any permutation of the indices $i_1, i_2, \dots, i_l, j_1, j_2, \dots, j_k$ such that the vertices corresponding to
	$r_p, u_p$ 
	are in the same block of $\beta$, and
	\begin{equation} \label{gammarpup}
		\gamma_{r_p, u_p} \coloneqq
		\begin{cases}
			\delta_{r_p, u_p} & \text{if the vertices corresponding to } r_p, u_p \text{ are in different rows of } d_\beta \\
			\epsilon_{r_p, u_p} & \text{if the vertices corresponding to } r_p, u_p \text{ are in the same row of } d_\beta
    		\end{cases}
	\end{equation}
	Here, $\epsilon_{r_p, u_p}$ is given by
\begin{equation} \label{epsilondef1}
	\epsilon_{\alpha, \beta} = \epsilon_{{\alpha'}, {\beta'}} = 0
\end{equation}
\begin{equation} \label{epsilondef2}
	\epsilon_{\alpha, {\beta'}} = - \epsilon_{{\alpha'}, {\beta}} = \delta_{\alpha, \beta}
\end{equation}
	\end{defn}

	\begin{defn}[$H_\alpha$ given in Theorem \ref{spanningsetSO(n)}]
		Suppose that $d_\alpha$ is a $(l+k)\backslash n$--diagram.
		Then $H_\alpha$ is defined as follows.

	Associate 
the indices $i_1, i_2, \dots, i_l$ with the vertices in the top row of $d_\alpha$, and $j_1, j_2, \dots, j_k$ with the vertices in the bottom row of $d_\alpha$.
	Suppose that there are $s$ free vertices in the top row. Then there are $n-s$ free vertices in the bottom row.
	Relabel the $s$ free indices in the top row (from left to right) by 
	$t_1, \dots, t_s$, and the $n-s$ free indices in the bottom row (from left to right) by $b_1, \dots, b_{n-s}$. 

	Then, define
		$
		\chi
			\left(\begin{smallmatrix} 
				1 & 2 & \cdots & s & s+1 & \cdots & n\\
				t_1 & t_2 & \cdots & t_s & b_1 & \cdots & b_{n-s}
			\end{smallmatrix}\right)
		$
		as follows: it is 
		$0$ if the elements $t_1, \dots, t_s, b_1, \dots, b_{n-s}$ are not distinct, otherwise, it is
		$	
		\sgn
			\left(\begin{smallmatrix} 
				1 & 2 & \cdots & s & s+1 & \cdots & n\\
				t_1 & t_2 & \cdots & t_s & b_1 & \cdots & b_{n-s}
			\end{smallmatrix}\right)
		$,
		considered as a permutation of $[n]$.

	As a result, for any $n \in \mathbb{Z}_{\geq 1}$, we have that
	\begin{equation} \label{SO(n)Halpha}
		H_\alpha
		\coloneqq
		\sum_{I \in [n]^l, J \in [n]^k} 
		\chi
			\left(\begin{smallmatrix} 
				1 & 2 & \cdots & s & s+1 & \cdots & n\\
				t_1 & t_2 & \cdots & t_s & b_1 & \cdots & b_{n-s}
			\end{smallmatrix}\right)
		\delta_{r_1, u_1}
		\delta_{r_2, u_2}
		\dots
		\delta_{r_{\frac{l+k-n}{2}}, u_{\frac{l+k-n}{2}}}
		E_{I,J}
	\end{equation}
	Here, $r_1, u_1, \dots, r_{\frac{l+k-n}{2}}, u_{\frac{l+k-n}{2}}$
	is any permutation of the indices 
	\begin{equation}	
	\{i_1, \dots, i_l, j_1, \dots, j_k\} \backslash \{t_1, \dots, t_s, b_1, \dots, b_{n-s}\}
	\end{equation}
	such that the vertices corresponding to $r_p, u_p$ are in the same block of $\alpha$. 
	\end{defn}

	\section{\textsc{PlanarMult} Procedure}

	We show how the matrix multiplication is performed in the \textsc{PlanarMult} procedure for each of the groups.

	\subsection{Orthogonal Group $O(n)$} \label{planarmultimpon}

	We take as input the planar $(k,l)$--Brauer diagram $d_\beta$ that is the output of \textsc{Factor}, and a vector $v \in (\mathbb{R}^{n})^{\otimes k}$ that is the output of \textsc{Permute}, as per Algorithm \ref{alg1}.
	
	Let $t$ be the number of pairs that are solely in the top row of $d_\beta$,
	let $d$ be the number of pairs that are in different rows of $d_\beta$, and
	let $b$ be the number of pairs that are solely in the bottom row of $d_\beta$.

	Then it is clear that
	\begin{equation}
		2t + d = l \quad \text{and} \quad 2b + d = k
	\end{equation}
	and so
	\begin{equation}
		2t + 2d + 2b = l+k 
	\end{equation}

	Given how \textsc{Factor} constructs the planar $(k,l)$--Brauer diagram $d_\beta$, the Brauer partition $\beta$ corresponding to $d_\beta$ will be of the form
	\begin{equation} \label{brauerpartfactor}
		\left(\bigcup_{i = 1}^{t} T_i \right)
		\bigcup
		\left(\bigcup_{i = 1}^{d} D_i \right)
		\bigcup
		\left(\bigcup_{i = 1}^{b} B_i \right)
	\end{equation}
	where we have used $T_i$ to refer to a top row pair, $D_i$ to refer to a different row pair, and $B_i$ to refer to a bottom row pair.
	In particular, we have that
	\begin{itemize}
		\item $T_i = \{2i-1, 2i\}$ for all $i = 1 \rightarrow t$
		\item $D_i = \{l+i-d, l+i\}$ for all $i = 1 \rightarrow d$, and
		\item $B_i = \{l+k-2b+2i-1, l+k-2b+2i\}$ for all $i = 1 \rightarrow b$.
	\end{itemize}

	As stated in the main text, we take $d_\beta$ and express it as a tensor product of three types of Brauer diagrams.
	The right-most type is itself a tensor product over diagrams corresponding to the $B_i$; the middle type is a single diagram corresponding to $\bigcup_{i = 1}^{d} D_i$; and the left-most type is itself a tensor product over diagrams corresponding to the $T_i$.

	We now apply the monoidal functor $\Phi$ to this tensor product decomposition of diagrams, which returns a Kronecker product of matrices.
	We perform the matrix multiplication by applying the matrices "right--to--left, diagram--by--diagram", as follows.

	\textbf{Step 1: Apply each matrix corresponding to a bottom row pair diagram, one by one, starting from the one that corresponds to $B_b$ and ending with the one that corresponds to $B_1$.}

	Suppose that we are performing the part of the matrix multiplication that corresponds to $B_i$, for some $i = 1 \rightarrow b$.
	The input will be a vector
	$w \in (\mathbb{R}^{n})^{\otimes k - 2(b-i)}$.

	We can express $w$ in the standard basis of $\mathbb{R}^{n}$ as
	\begin{equation}
		w = \sum_{L \in [n]^{k-2(b-i)}} w_Le_L
	\end{equation}

	This will be mapped to the vector
	$r \in (\mathbb{R}^{n})^{\otimes k - 2(b-i) - 2}$,
	where $r$ is of the form
	\begin{equation}
		r = \sum_{M \in [n]^{k-2(b-i)-2}} r_Me_M
	\end{equation}
	and
	\begin{equation} \label{rMcoeff}
		r_M = \sum_{j=1}^{n} w_{M,j,j}
	\end{equation}
	At the end of this process, we obtain a vector in 
	$(\mathbb{R}^{n})^{\otimes k - 2b}$.
	
	Note that the matrices corresponding to bottom row pairs are merely performing indexing and summation operations, that is, ultimately, tensor contractions.

	\textbf{Step 2: Now apply the matrix corresponding to the middle diagram, that is, to the set $\bigcup_{i = 1}^{d} D_i$.}
	
	The input will be a vector
	$w \in (\mathbb{R}^{n})^{\otimes k - 2b}$.
	
	Expressing $w$ in the standard basis of $\mathbb{R}^{n}$ as
	\begin{equation}
		w = \sum_{L \in [n]^{k-2b}} w_Le_L
	\end{equation}
	the multiplication of the matrix merely returns $w$ itself!
	
	These operations are called transfer operations, and, for the orthogonal group, they are simply the identity map!

	\textbf{Step 3: Finally, apply each matrix corresponding to a top row pair diagram, one by one, starting from the one that corresponds to $T_t$ and ending with the one that corresponds to $T_1$.}
	
	Suppose that we are performing the part of the matrix multiplication that corresponds to $T_i$, for some $i = 1 \rightarrow t$.

	Then we begin with a vector  
	$w \in (\mathbb{R}^{n})^{\otimes k - 2b + 2(t - i)}$ 
	that is of the form
	\begin{equation}
		w = \sum_{J \in [n]^{t-i}}\sum_{L \in [n]^{k-2b}} v_L \left(e_{j_1} \otimes e_{j_1} \otimes \dots \otimes e_{j_{t-i}} \otimes e_{j_{t-i}} \otimes e_L\right)
	\end{equation}
	where $v_L$ is the coefficient of $e_L$ appearing in the vector at the end of step 2.
	
	This will be mapped to the vector
	$r \in (\mathbb{R}^{n})^{\otimes k - 2b+2(t-i) + 2}$,
	where $r$ is of the form
	\begin{equation}
		r = \sum_{m \in [n]}\sum_{J \in [n]^{t-i}}\sum_{L \in [n]^{k-2b+2(t-i)}} v_L \left(e_m \otimes e_m \otimes e_{j_1} \otimes e_{j_1} \otimes \dots \otimes e_{j_{t-i}} \otimes e_{j_{t-i}} \otimes e_L\right)
	\end{equation}
	At the end of this process, we obtain a vector in 
	$(\mathbb{R}^{n})^{\otimes l}$, since $k - 2b + 2t = l$, which is of the form
	\begin{equation}
		\sum_{J \in [n]^t}\sum_{L \in [n]^{k-2b}} v_L \left(e_{j_1} \otimes e_{j_1} \otimes \dots \otimes e_{j_t} \otimes e_{j_t} \otimes e_L\right)
	\end{equation}
	This is the vector that is returned by \textsc{PlanarMult} for the orthogonal group $O(n)$.

	We give an example in Appendix \ref{MatrixMultExamplesO(n)}.

	\subsection{Symplectic Group $Sp(n)$}

	Again, we take as input the planar $(k,l)$--Brauer diagram $d_\beta$ that is the output of \textsc{Factor}, and a vector $v \in (\mathbb{R}^{n})^{\otimes k}$ that is the output of \textsc{Permute}, as per Algorithm \ref{alg1}.

	The tensor product decomposition of $d_\beta$ is the same as for the orthogonal group $O(n)$; in particular, the Brauer partition $\beta$ corresponding to $d_\beta$ is of the form (\ref{brauerpartfactor}).

	The important difference here is that we apply the monoidal functor $X$, instead of $\Phi$, to this tensor product decomposition of diagrams, which returns a different Kronecker product of matrices.
	Again, we perform the same steps for the matrix multiplication, but this time the vectors returned at each stage are different.

	\textbf{Step 1: Apply each matrix corresponding to a bottom row pair diagram, one by one, starting from the one that corresponds to $B_b$ and ending with the one that corresponds to $B_1$.}

	Suppose that we are performing the part of the matrix multiplication that corresponds to $B_i$, for some $i = 1 \rightarrow b$.
	The input will be a vector
	$w \in (\mathbb{R}^{n})^{\otimes k - 2(b-i)}$.

	We can express $w$ in the standard basis of $\mathbb{R}^{n}$ as
	\begin{equation}
		w = \sum_{L \in [n]^{k-2(b-i)}} w_Le_L
	\end{equation}

	This will be mapped to the vector
	$r \in (\mathbb{R}^{n})^{\otimes k - 2(b-i) - 2}$ 
	where $r$ is of the form
	\begin{equation}
		r = \sum_{M \in [n]^{k-2(b-i)-2}} r_Me_M
	\end{equation}
	where
	\begin{equation} 
		r_M = \sum_{j_{k-2(b-i)-1}, j_{k-2(b-i)} = 1}^{n} 
		\epsilon_{j_{k-2(b-i)-1},j_{k-2(b-i)}}
		w_{M,j_{k-2(b-i)-1},j_{k-2(b-i)}}
	\end{equation}
	Recall that $\epsilon_{j_{k-2(b-i)-1},j_{k-2(b-i)}}$ was defined in (\ref{epsilondef1}) and (\ref{epsilondef2}). 

	At the end of this process, we obtain a vector in 
	$(\mathbb{R}^{n})^{\otimes k - 2b}$.
	
	As before, these matrices corresponding to bottom row pairs are merely copying arrays and performing summations, that is, tensor contractions.

	\textbf{Step 2: Now apply the matrix corresponding to the middle diagram, that is, to the set $\bigcup_{i = 1}^{d} D_i$.}

	This will be exactly the same as for the orthogonal group, by the definition of $\gamma_{r_p, u_p}$ given in (\ref{gammarpup}).

	\textbf{Step 3: Finally, apply each matrix corresponding to a top row pair diagram, one by one, starting from the one that corresponds to $T_t$ and ending with the one that corresponds to $T_1$.}

	Suppose that we are performing the part of the matrix multiplication that corresponds to $T_i$, for some $i = 1 \rightarrow t$.

	Then we begin with a vector  
	$w \in (\mathbb{R}^{n})^{\otimes k - 2b + 2(t - i)}$ 
	that is of the form
	\begin{equation}
		w = 
		\sum_{J \in [n]^{2(t-i)}}\sum_{L \in [n]^{k-2b}} 
		\epsilon_J
		v_L 
		\left(
		e_J \otimes e_L\right)
	\end{equation}
	where 
	\begin{equation}
		\epsilon_J
		\coloneqq
		\epsilon_{j_1, j_2}
		\dots
		\epsilon_{j_{2(t-i)-1}, j_{2(t-i)}}
	\end{equation}
	and 
	where $v_L$ is the coefficient of $e_L$ appearing in the vector at the end of step 2.
	
	This will be mapped to the vector
	$r \in (\mathbb{R}^{n})^{\otimes k - 2b+2(t-i) + 2}$,
	where $r$ is of the form
	\begin{equation}
		r = 
		\sum_{M \in [n]^2}
		\sum_{J \in [n]^{2(t-i)}}\sum_{L \in [n]^{k-2b}} 
		\epsilon_{M,J}
		v_L 
		\left(
		e_M \otimes e_J	\otimes e_L\right)
	\end{equation}
	where
	\begin{equation}
		\epsilon_{M,J}
		\coloneqq
		\epsilon_{m_1, m_2}
		\epsilon_{j_1, j_2}
		\dots
		\epsilon_{j_{2(t-i)-1}, j_{2(t-i)}}
	\end{equation}
	At the end of this process, we obtain a vector in 
	$(\mathbb{R}^{n})^{\otimes l}$, since $k - 2b + 2t = l$, which is of the form
	\begin{equation}
		\sum_{J \in [n]^{2t}}\sum_{L \in [n]^{k-2b}} 
		\epsilon_{J}
		v_L \left(
		e_J \otimes e_L\right)
	\end{equation}
	where $\epsilon_{J}$ is redefined to be
	\begin{equation}
		\epsilon_{j_1, j_2}
		\dots
		\epsilon_{j_{2t-1}, j_{2t}}
	\end{equation}
	This is the vector that is returned by \textsc{PlanarMult} for the symplectic group $Sp(n)$.

	We give an example in Appendix \ref{MatrixMultExamplesSp(n)}.

	\subsection{Special Orthogonal Group $SO(n)$} \label{planarmultimpSOn}

	We stated in the main text that if
	the input diagram is a planar $(k,l)$--Brauer diagram $d_\beta$, 
	then the procedure works in exactly the same way as for $O(n)$.

	Hence, we consider the case where we take as input the planar $(l+k) \backslash n$--diagram $d_\alpha$ that
	is the output of \textsc{Factor}, and a vector $v \in (\mathbb{R}^{n})^{\otimes k}$ that is the output of \textsc{Permute}, as per Algorithm \ref{alg1}.

	We need new subsets (and notation) to consider the impact of the free vertices in the diagram $d_\alpha$.

	As before,
	let $t$ be the number of pairs that are solely in the top row of $d_\alpha$,
	let $d$ be the number of pairs that are in different rows of $d_\alpha$, and
	let $b$ be the number of pairs that are solely in the bottom row of $d_\alpha$. 
	Now, let $s$ be the number of free vertices in the top row of $d_\alpha$. Hence there are $n-s$ free vertices in the bottom row of $d_\alpha$.

	Then it is clear that
	\begin{equation} \label{SO(n)restrictions}
		2t + d + s = l \quad \text{and} \quad 2b + d + n - s = k
	\end{equation}
	and so
	\begin{equation}
		2t + 2d + 2b + n = l+k 
	\end{equation}

	Given how \textsc{Factor} constructs the planar $(l+k) \backslash n$--diagram $d_\alpha$, the set partition $\alpha$ corresponding to $d_\alpha$ will be of the form
	\begin{equation}
		\left(\bigcup_{i = 1}^{t} T_i \right)
		\bigcup
		\left(\bigcup_{i = 1}^{d} D_i \right)
		\bigcup
		\left(\bigcup_{i = 1}^{s} TF_i \right)
		\bigcup
		\left(\bigcup_{i = 1}^{b} B_i \right)
		\bigcup
		\left(\bigcup_{i = 1}^{n-s} BF_i \right)
	\end{equation}
	where we have used $T_i$ to refer to a top row pair, $D_i$ to refer to a different row pair, 
	$TF_i$ to refer to a top row free vertex,
	$B_i$ to refer to a bottom row pair, and
	$BF_i$ to refer to a bottom row free vertex.
	In particular, we have that
	\begin{itemize}
		\item $T_i = \{2i-1, 2i\}$ for all $i = 1 \rightarrow t$,
		\item $D_i = \{l+i-d-s, l+i\}$ for all $i = 1 \rightarrow d$,
		\item $TF_i = \{l-s+i\}$ for all $i = 1 \rightarrow s$, 
		\item $B_i = \{l+d+2i-1, l+d+2i\}$ for all $i = 1 \rightarrow b$, and
		\item $BF_i = \{l+d+2b+i\}$ for all $i = 1 \rightarrow n-s$.
	\end{itemize}

	As stated in the main text, we take $d_\alpha$ and express it as a tensor product of four types of set partition diagrams.

	The right-most type is a diagram consisting of all the free vertices, hence it corresponds to
	$	
		\left(\bigcup_{i = 1}^{s} TF_i \right)
		\bigcup
		\left(\bigcup_{i = 1}^{n-s} BF_i \right)
	$.
	The type to its left is itself a tensor product over diagrams corresponding to the $B_i$; the next is a single diagram corresponding to
	$\left(\bigcup_{i = 1}^{d} D_i\right)$; and, finally, the left-most type is itself a tensor product over diagrams corresponding to the $T_i$.

	We now apply the monoidal functor $\Psi$ to this tensor product decomposition of diagrams, which returns a Kronecker product of matrices.
	We perform the matrix multiplication by applying the matrices "right--to--left, diagram--by--diagram", as follows.

	\textbf{Step 1: Apply the matrix that corresponds to the free vertices, that is, to the set 
	$	
		\left(\bigcup_{i = 1}^{s} TF_i \right)
		\bigcup
		\left(\bigcup_{i = 1}^{n-s} BF_i \right)
	$.}

	The input will be a vector
	$v \in (\mathbb{R}^{n})^{\otimes k}$.

	As $k = 2b + d + (n-s)$, we can express $v$ 
	in the standard basis of $\mathbb{R}^{n}$ as
	\begin{equation} \label{SOnStep1,1}
		v = 
		\sum_{J \in [n]^{2b + d}} 
		\sum_{B \in [n]^{n-s}} 
		v_{J,B}e_{J,B}
	\end{equation}
	This will be mapped to the vector $r \in (\mathbb{R}^{n})^{\otimes 2b+d+s}$, where $r$ is of the form
	\begin{equation} \label{SOnStep1,2}
		w =
		\sum_{J \in [n]^{2b + d}} 
		\sum_{T \in [n]^{s}} 
		w_{J,T}e_{J,T}
	\end{equation}
	where
	\begin{equation} \label{SOnStep1,3}
		w_{J,T} =
		\sum_{B \in [n]^{n-s}} 
		v_{J,B}
		\;
		\chi
			\left(\begin{smallmatrix} 
				1 & 2 & \cdots & s & s+1 & \cdots & n\\
				t_1 & t_2 & \cdots & t_s & b_1 & \cdots & b_{n-s}
			\end{smallmatrix}\right)
	\end{equation}


	\textbf{Step 2: Apply each matrix corresponding to a bottom row pair diagram, one by one, starting from the one that corresponds to $B_b$ and ending with the one that corresponds to $B_1$.}

	This is exactly the same as Step 1 for the orthogonal group.
	We obtain a vector $y \in (\mathbb{R}^{n})^{\otimes d+s}$

	\textbf{Step 3: Now apply the matrix corresponding to the middle diagram, that is, to the set 
	$\left(\bigcup_{i = 1}^{d} D_i\right)$.}

	This is exactly the same as Step 2 for the orthogonal group.
	We obtain a vector $r \coloneqq y \in (\mathbb{R}^{n})^{\otimes d+s}$

	\textbf{Step 4: Finally, apply each matrix corresponding to a top row pair diagram, one by one, starting from the one that corresponds to $T_t$ and ending with the one that corresponds to $T_1$.}
	
	This is exactly the same as Step 3 for the orthogonal group.
	We obtain a vector $z \in (\mathbb{R}^{n})^{\otimes 2t+d+s}$.
	As $2t+d+s = l$ by (\ref{SO(n)restrictions}),
	we have that $z \in (\mathbb{R}^{n})^{\otimes l}$, as required.
	
	We give an example in Appendix \ref{MatrixMultExamplesSO(n)}.

	\section{Computational Cost Analysis of \textsc{MatrixMult}}

	We compare how the implementation for the \textsc{MatrixMult} algorithm performs in terms of computational cost relative to a naive implementation for performing the matrix multiplication.

	Clearly, the naive implementation of multiplying a matrix in 
	$\Hom_{G}((\mathbb{R}^{n})^{\otimes k}, (\mathbb{R}^{n})^{\otimes l})$
	with a vector
	$v \in (\mathbb{R}^{n})^{\otimes k}$,
	expressed in the standard basis of $\mathbb{R}^{n}$,
	consists of
	$n^{l+k}$ multiplications
	and
	$n^{l}(n^{k}-1)$ additions,
	for an overall $O(n^{l+k})$ time complexity.

	We consider the implementation for each group $G$.
	Note that in our analysis, we are viewing memory operations, such as permuting basis vectors and making copies of coefficients, as having no cost.
	Hence, in particular, we view the procedures \textsc{Factor} and \textsc{Permute} as having no cost; consequently, by Algorithm \ref{alg1}, it is enough to consider the computational cost of \textsc{PlanarMult} only.
	We consider the cost for each group in turn.

	\subsection{Orthogonal Group $O(n)$}

	We look at each step of the implementation that was given in Appendix 
	\ref{planarmultimpon}.

	\textbf{Step 1:} For the part of the matrix multiplication that corresponds to $B_i$, for some $i = 1 \rightarrow b$, we map a vector in 
	$(\mathbb{R}^{n})^{\otimes k - 2(b-i)}$ 
	to a vector in
	$(\mathbb{R}^{n})^{\otimes k - 2(b-i) - 2}$.

	Since the matrix corresponds to a bottom row pair that is connected, for each tuple $M$ of indices in the output coefficient $r_M$, as in (\ref{rMcoeff}),
	there are only $n$ terms to multiply (an improvement over $n^2$), and consequently only $n-1$ additions.

	Hence, in total, there are
	\begin{equation}
		\sum_{i = 1}^{b}
		n^{k - 2(b-i) - 1}
	\end{equation}
	multiplications and
	\begin{equation}
		\sum_{i = 1}^{b}
		n^{k - 2(b-i) -2} \cdot (n-1)
	\end{equation}
	additions, for an overall time complexity of $O(n^{k-1})$.

	\textbf{Step 2:} This corresponds to the identity transformation, hence there is no cost, as we do not need to perform this operation.

	\textbf{Step 3:} Here we are copying arrays, hence there is no cost.

	Consequently, we have reduced the overall time complexity from 
	$O(n^{l+k})$ to 
	$O(n^{k-1})$.

	\subsection{Symplectic Group $Sp(n)$}

	The analysis is exactly the same as for the orthgonal group.

	\subsection{Special Orthogonal Group $SO(n)$}

	For a matrix corresponding to a $(k,l)$--Brauer diagram, the analysis is the same as for the orthgonal group.
	For a matrix corresponding to a $(l+k) \backslash n$--diagram, we look at each step of the implementation that was given in Appendix
	\ref{planarmultimpSOn}.

	\textbf{Step 1:} 
	Recall that this step maps a vector in 
	$(\mathbb{R}^{n})^{\otimes k}$ 
	to a vector in
	$(\mathbb{R}^{n})^{\otimes 2b+d+s}$.

	To obtain the best time complexity, we need to look at the tuples 
	$J \in [n]^{2b+d}, T \in [n]^{s}$ and $B \in [n]^{n-s}$
	appearing in
	(\ref{SOnStep1,1}),
	(\ref{SOnStep1,2}), and
	(\ref{SOnStep1,3}).

	Indeed, for each tuple $J$, we first need to consider how many tuples $T$ come with pairwise different entries in $[n]$.
	The number of such tuples is $\frac{n!}{(n-s)!}$.
	Consequently, we only need to perform multiplications and additions for entries with such indices $(J,T)$.

	Given the definition of $\chi$, for each choice of $T$, there are $(n-s)!$ choices of $B$ that give a non-zero value of $\chi$.
	Hence, for each tuple $J$, there are
	\begin{equation}
		\frac{n!}{(n-s)!}(n-s)! = n!
	\end{equation}
	multiplications and
	\begin{equation}
		\frac{n!}{(n-s)! - 1}(n-s)! 
	\end{equation}
	additions, making for an overall time complexity of
	$O(n^{k - (n-s)}n!)$ for this step, since $2b+d = k - (n-s)$, by (\ref{SO(n)restrictions}).

	\textbf{Step 2:} By the operations performed in this step, we can immediately apply the analysis of Step 1 for the orthogonal group.
	
	Since the part of the matrix multiplication that corresponds to $B_i$, for some $i = 1 \rightarrow b$, maps a vector in	
	$(\mathbb{R}^{n})^{\otimes d+s+2i}$ 
	to a vector in
	$(\mathbb{R}^{n})^{\otimes d+s+2i-2}$, we have that
	the overall time complexity is
	$O(n^{k+s-(n-s)-1})$, since
	$2b + d + s = k + s - (n-s)$,
	by (\ref{SO(n)restrictions}).

	\textbf{Steps 3, 4:} These steps correspond to steps 2, 3 for the orthgonal group, hence they have no cost.
	
	Consequently, we have reduced the overall time complexity from
	$O(n^{l+k})$
	to
	\begin{equation}	
		O(n^{k - (n-s)}(n! + n^{s-1}))
	\end{equation}

	\section{Examples of \textsc{MatrixMult}} 
	\label{MatrixMultExamples}

	\subsection{Orthogonal Group $O(n)$}
	\label{MatrixMultExamplesO(n)}

	Suppose that we wish to perform the multiplication of $E_\beta$
	by $v \in (\mathbb{R}^{n})^{\otimes 5}$, where
	$E_\beta$ corresponds to the $(5,5)$--Brauer diagram $d_\beta$ given in Figure \ref{symmfactoringOn}a),
	and $v$ is given by
		\begin{equation}
			\sum_{L \in [n]^5} v_Le_L
		\end{equation}

	We know that $E_\beta$ is a matrix in
	$\Hom_{O(n)}((\mathbb{R}^{n})^{\otimes 5}, (\mathbb{R}^{n})^{\otimes 5})$.

	First, we apply the procedure \textsc{Factor}, which returns the three diagrams given in Figure \ref{symmfactoringOn}a).
	The first diagram corresponds to the permutation $(1524)$ in $S_5$, hence, the result of \textsc{Permute}$(v, (1524))$ is the vector
		\begin{equation}
			\sum_{L \in [n]^5} 
			v_{l_1, l_2, l_3, l_4, l_5}
			\left(
				e_{l_5} \otimes 
				e_{l_4} \otimes 
				e_{l_3} \otimes
				e_{l_1} \otimes
				e_{l_2} 
			\right)
		\end{equation}
	Now we apply \textsc{PlanarMult} with the decomposition given in Figure \ref{tensorproddecompOn}a).

	Step 1: Apply the matrices that correspond to the bottom row pairs.

	We obtain the vector
		\begin{equation}
			w = 
			\sum_{l_5, l_4, l_3 \in [n]} 
			w_{l_5, l_4, l_3}
			\left(
				e_{l_5} \otimes 
				e_{l_4} \otimes 
				e_{l_3}
			\right)
		\end{equation}
	where
		\begin{equation}
			w_{l_5, l_4, l_3}
			=
			\sum_{j \in [n]}
			v_{j, j, l_3, l_4, l_5}
		\end{equation}

	Step 2: Apply the matrices that correspond to the middle diagram.

	As the transfer operations are the identity mapping, we get $w$.	
	
	Step 3: Apply the matrices that correspond to the top row.

	We obtain the vector
		\begin{equation}
			z =
			\sum_{m \in [n]}	
			\sum_{l_5, l_4, l_3 \in [n]} 
			w_{l_5, l_4, l_3}
			\left(
				e_{m} \otimes e_{m} \otimes
				e_{l_5} \otimes 
				e_{l_4} \otimes 
				e_{l_3}
			\right)
		\end{equation}
	Substituting in, we get that
		\begin{equation}
			z =
			\sum_{m \in [n]}	
			\sum_{l_5, l_4, l_3 \in [n]} 
			\sum_{j \in [n]}
			v_{j, j, l_3, l_4, l_5}
			\left(
				e_{m} \otimes 
				e_{m} \otimes
				e_{l_5} \otimes 
				e_{l_4} \otimes 
				e_{l_3}
			\right)
		\end{equation}
	Finally, as the third diagram returned from \textsc{Factor} 
	corresponds to the permutation $(1342)$ in $S_5$,
	we perform \textsc{Permute}$(z, (1342))$,
	which returns the vector
		\begin{equation}
			\sum_{m \in [n]}	
			\sum_{l_5, l_4, l_3 \in [n]} 
			\sum_{j \in [n]}
			v_{j, j, l_3, l_4, l_5}
			\left(
				e_{l_5} \otimes 
				e_{m} \otimes 
				e_{l_4} \otimes 
				e_{m} \otimes
				e_{l_3}
			\right)
		\end{equation}
	This is the vector that is returned by \textsc{MatrixMult}.	
	
	\subsection{Symplectic Group $Sp(n)$}
	\label{MatrixMultExamplesSp(n)}

	Suppose instead that we wish to perform the multiplication of $F_\beta$ 
	by $v \in (\mathbb{R}^{n})^{\otimes 5}$,
	for the same $(5,5)$--Brauer diagram $d_\beta$ given in Figure \ref{symmfactoringOn}a),
	and $v$ is given by
		\begin{equation}
			\sum_{L \in [n]^5} v_Le_L
		\end{equation}

	We know that $F_\beta$ is a matrix in
	$\Hom_{Sp(n)}((\mathbb{R}^{n})^{\otimes 5}, (\mathbb{R}^{n})^{\otimes 5})$.

	The \textsc{Factor} and \textsc{Permute} steps are the same as for the previous example, hence, prior to the \textsc{PlanarMult} step, we have the vector
		\begin{equation}
			\sum_{L \in [n]^5} 
			v_{l_1, l_2, l_3, l_4, l_5}
			\left(
				e_{l_5} \otimes 
				e_{l_4} \otimes 
				e_{l_3} \otimes
				e_{l_1} \otimes
				e_{l_2} 
			\right)
		\end{equation}
	We now apply \textsc{PlanarMult} with the decomposition given in Figure \ref{tensorproddecompOn}a).

	Step 1: Apply the matrices that correspond to the bottom row pairs.

	We obtain the vector
		\begin{equation}
			w = 
			\sum_{l_5, l_4, l_3 \in [n]} 
			w_{l_5, l_4, l_3}
			\left(
				e_{l_5} \otimes 
				e_{l_4} \otimes 
				e_{l_3}
			\right)
		\end{equation}
	where
		\begin{equation}
			w_{l_5, l_4, l_3}
			=
			\sum_{j_1, j_2 \in [n]}
			\epsilon_{j_1, j_2}
			v_{j_1, j_2, l_3, l_4, l_5}
		\end{equation}

	Step 2: Apply the matrices that correspond to the middle diagram.

	As the transfer operations are the identity mapping, we get $w$.	
	
	Step 3: Apply the matrices that correspond to the top row.

	We obtain the vector
		\begin{equation}
			z =
			\sum_{m_1, m_2 \in [n]}	
			\sum_{l_5, l_4, l_3 \in [n]} 
			\epsilon_{m_1, m_2}
			w_{l_5, l_4, l_3}
			\left(
				e_{m_1} \otimes e_{m_2} \otimes
				e_{l_5} \otimes 
				e_{l_4} \otimes 
				e_{l_3}
			\right)
		\end{equation}
	Substituting in, we get that
		\begin{equation}
			z =
			\sum_{m_1, m_2 \in [n]}	
			\sum_{l_5, l_4, l_3 \in [n]} 
			\sum_{j_1, j_2 \in [n]}
			\epsilon_{m_1, m_2}
			\epsilon_{j_1, j_2}
			v_{j_1, j_2, l_3, l_4, l_5}
			\left(
				e_{m_1} \otimes 
				e_{m_2} \otimes
				e_{l_5} \otimes 
				e_{l_4} \otimes 
				e_{l_3}
			\right)
		\end{equation}
	Finally, as the third diagram returned from \textsc{Factor} 
	corresponds to the permutation $(1342)$ in $S_5$,
	we perform \textsc{Permute}$(z, (1342))$,
	which returns the vector
		\begin{equation}
			\sum_{m_1, m_2 \in [n]}	
			\sum_{l_5, l_4, l_3 \in [n]} 
			\sum_{j_1, j_2 \in [n]}
			\epsilon_{m_1, m_2}
			\epsilon_{j_1, j_2}
			v_{j_1, j_2, l_3, l_4, l_5}
			\left(
				e_{l_5} \otimes 
				e_{m_1} \otimes 
				e_{l_4} \otimes 
				e_{m_2} \otimes
				e_{l_3}
			\right)
		\end{equation}
	This is the vector that is returned by \textsc{MatrixMult}.	

	\begin{figure}
  		\centering
		\scalebox{0.33}{\input{planarmultSOn.tikz}}
		\caption{We show how matrix multiplication is implemented in \textsc{PlanarMult} for $SO(n)$ (here $n=3$) using the tensor product decomposition of the planar 
		$(4+5) \backslash 3$--diagram 
		given in Figure \ref{tensorproddecompOn}b)
		as an example.
		We perform the matrix multiplication as follows:
		first, we deform the entire tensor product decomposition diagram by pulling each individual diagram up one level higher than the previous one, going from right--to-left, and then we apply the functor $\Psi$ at each level. 
Note, in particular, that we need to attach identity strings to the free vertices appearing in the top row.
		Finally, we perform matrix multiplication at each level to obtain the final output vector.}
		\label{planarmultSOn}
	\end{figure}

	\subsection{Special Orthogonal Group $SO(n)$}
	\label{MatrixMultExamplesSO(n)}

	Suppose that we wish to perform the multiplication of $H_\alpha$
	by $v \in (\mathbb{R}^{3})^{\otimes 5}$, where
	$H_\alpha$ corresponds to the $(4+5) \backslash 3$--diagram $d_\alpha$ given in Figure \ref{symmfactoringOn}b),
	and $v$ is given by
		\begin{equation}
			\sum_{L \in [3]^5} v_Le_L
		\end{equation}
	when expressed in the standard basis of $\mathbb{R}^{3}$.
	
	We know that $H_\alpha$ is a matrix in
	$\Hom_{SO(3)}((\mathbb{R}^{3})^{\otimes 5}, (\mathbb{R}^{3})^{\otimes 4})$.

	First, we apply the procedure \textsc{Factor}, which returns the three diagrams given in Figure \ref{symmfactoringOn}b).
	The first diagram corresponds to the permutation $(13524)$ in $S_5$, hence, the result of \textsc{Permute}$(v, (13524))$ is the vector
		\begin{equation}
			\sum_{L \in [3]^5} 
			v_{l_1, l_2, l_3, l_4, l_5}
			\left(
				e_{l_3} \otimes 
				e_{l_4} \otimes 
				e_{l_5} \otimes
				e_{l_1} \otimes
				e_{l_2} 
			\right)
		\end{equation}
	We now apply \textsc{PlanarMult} with the decomposition given in Figure \ref{tensorproddecompOn}b).

	Step 1: Apply the matrices that correspond to the free vertices.
	
	We obtain the vector
		\begin{equation}
			w = 
			\sum_{l_3, l_4, l_5, t_1 \in [3]} 
			w_{l_3, l_4, l_5, t_1}
			\left(
				e_{l_3} \otimes 
				e_{l_4} \otimes 
				e_{l_5} \otimes 
				e_{t_1}
			\right)
		\end{equation}
	where
		\begin{equation}
			w_{l_3, l_4, l_5, t_1}
			=
			\sum_{l_1, l_2 \in [3]}
			v_{l_1, l_2, l_3, l_4, l_5}
				\;
		\chi
			\left(\begin{smallmatrix} 
				1 & 2 & 3\\
				t_1 & l_1 & l_2
			\end{smallmatrix}\right)
		\end{equation}

	Step 2: Apply the matrices that correspond to the bottom row pairs.

	We obtain the vector
		\begin{equation}
			y = 
			\sum_{l_3, t_1 \in [3]} 
			y_{l_3, t_1}
			\left(
				e_{l_3} \otimes 
				e_{t_1}
			\right)
		\end{equation}
	where
		\begin{equation}
			y_{l_3, t_1}
			=
			\sum_{j \in [3]}
			w_{l_3, j, j, t_1}
		\end{equation}
	
	Step 3: Apply the matrices that correspond to the different row pairs.

	Here, we get that $r = y$, as the transfer operations correspond to the identity.

	Step 4: Apply the matrices that correspond to the top row pairs.

	We obtain the vector
		\begin{equation}
			z =
			\sum_{m \in [3]}	
			\sum_{l_3, t_1 \in [3]}
			y_{l_3, t_1}
			\left(	
			e_{m} \otimes e_{m} 
			\otimes e_{l_3} \otimes e_{t_1}
			\right)
		\end{equation}
	Substituting in, we get that
		\begin{equation}
			z =
			\sum_{m \in [3]}	
			\sum_{l_3, t_1 \in [3]}
			\sum_{j \in [3]}
			w_{l_3, j, j, t_1}
			\left(	
			e_{m} \otimes e_{m} 
			\otimes e_{l_3} \otimes e_{t_1}
			\right)
		\end{equation}
	and hence
		\begin{equation}
			z =
			\sum_{m \in [3]}	
			\sum_{l_3, t_1 \in [3]}
			\sum_{j \in [3]}
			\sum_{l_1, l_2 \in [3]}
			v_{l_1, l_2, l_3, j, j}
				\;
		\chi
			\left(\begin{smallmatrix} 
				1 & 2 & 3\\
				t_1 & l_1 & l_2
			\end{smallmatrix}\right)
			\left(	
			e_{m} \otimes e_{m} 
			\otimes e_{l_3} \otimes e_{t_1}
			\right)
		\end{equation}
	Finally, as the third diagram returned from \textsc{Factor} 
	corresponds to the permutation $(1432)$ in $S_4$,
	we perform \textsc{Permute}$(z, (1432))$
	which returns the vector
		\begin{equation}
			\sum_{m \in [3]}	
			\sum_{l_3, t_1 \in [3]}
			\sum_{j \in [3]}
			\sum_{l_1, l_2 \in [3]}
			v_{l_1, l_2, l_3, j, j}
				\;
		\chi
			\left(\begin{smallmatrix} 
				1 & 2 & 3\\
				t_1 & l_1 & l_2
			\end{smallmatrix}\right)
			\left(	
			e_{t_1} \otimes e_{m} \otimes e_{m} 
			\otimes e_{l_3} 
			\right)
		\end{equation}
	This is the vector that is returned by \textsc{MatrixMult}.	


	\section{Application to the Symmetric Group $S_n$}

	The method that we have presented in the main text can be adapted to construct an algorithm for multiplying any vector $v \in (\mathbb{R}^{n})^{\otimes k}$ by any matrix in 
	$\Hom_{S_n}((\mathbb{R}^{n})^{\otimes k}, (\mathbb{R}^{n})^{\otimes l})$.

	First, we introduce the following category.

	\begin{defn} \label{partitioncategory}
		The partition category $\mathcal{P}(n)$ is
		the category whose objects are the non--negative integers $\mathbb{N}_{\geq 0} = \{0, 1, 2, \dots \}$,
		and, for any pair of objects $k$ and $l$, the morphism space
		$\Hom_{\mathcal{P}(n)}(k,l)$ is defined to be the $\mathbb{R}$--linear span of the set of all $(k,l)$--partition diagrams.
	\end{defn}
	As before, this category comes with a vertical composition of morphisms, a tensor product operation on objects and morphisms, and a unit object.
	The vertical composition of morphisms can be found in \cite[Section 2.2]{pearcecrumpC}.
	The unit object is the object $0$.
	The tensor product operation on objects is given by the standard addition operation in $\mathbb{N}_{\geq 0}$.
	Finally, the tensor product operation is defined on diagrams (morphisms) as follows:
	if 
$d_{\pi_1}$ is a $(k,l)$--partition diagram
and
$d_{\pi_2}$ is a $(q,m)$--partition diagram,
then 
$d_{\pi_1} \otimes d_{\pi_2}$ 
is defined to be the $(k+q,l+m)$--partition diagram
obtained by horizontally placing
$d_{\pi_1}$
to the left of
$d_{\pi_2}$
without any overlapping of vertices.

$\mathcal{P}(n)$ is in fact a strict $\mathbb{R}$--linear monoidal category -- see \cite[Section 4.1]{pearcecrumpC} for more details.

Next, there is a basis for
	$\Hom_{S_n}((\mathbb{R}^{n})^{\otimes k}, (\mathbb{R}^{n})^{\otimes l})$
that is indexed by certain set partitions.
The basis that we use is called the diagram basis, which was discovered by \cite[Theorem 5.4]{godfrey}.
\begin{theorem}[Diagram Basis when $G = S_n$] 
	\cite[Theorem 5.4]{godfrey}
	\label{diagbasisSn}

	For any $k, l \in \mathbb{Z}_{\geq 0}$ and any
	$n \in \mathbb{Z}_{\geq 1}$, 	
	the set
	\begin{equation} \label{klSnSpanningSet}
		\{E_\pi \mid d_\pi \text{ is a } (k,l) \text{--partition diagram having at most } n \text{ blocks} \}
	\end{equation}
	is a basis for
	$\Hom_{S_n}((\mathbb{R}^{n})^{\otimes k}, (\mathbb{R}^{n})^{\otimes l})$
	in the standard basis of $\mathbb{R}^{n}$, where
		$E_\pi$ is defined as follows.

		Let $d_\pi$ be the $(k,l)$--partition diagram corresponding to $E_\pi$.

		Associate the indices $i_1, i_2, \dots, i_l$ with the vertices in the top row of $d_\pi$, and $j_1, j_2, \dots, j_k$ with the vertices in the bottom row of $d_\pi$.
	Then, if $S_\pi((I,J))$ is defined to be the set
	\begin{equation} \label{Snindexingset}
		\{(I,J) \in [n]^{l+k} \mid \text{if } x,y \text{ are in the same block of } \pi, \text{then } i_x = i_y \}
	\end{equation}
	(where we have momentarily replaced the elements of $J$ by $i_{l+m} \coloneqq j_m$ for all $m \in [k]$),
	 we have that
	\begin{equation} \label{mappeddiagbasisSn}
		E_\pi
		\coloneqq
		\sum_{I \in [n]^l, J \in [n]^k}
		\delta_{\pi, (I,J)}
		E_{I,J}
	\end{equation}
	where
	\begin{equation}
		\delta_{\pi, (I,J)}
		\coloneqq
		\begin{cases}
			1 & \text{if } (I,J) \in S_\pi((I,J)) \\
			0 & \text{otherwise}
		\end{cases}
	\end{equation}
	\end{theorem}
	The definition of the category $\mathcal{C}(G)$, given in Definition 
	\ref{catgroupreps}, extends to $G = S_n$.
	Hence, as before, we have the following functor, which was introduced in \cite[Theorem 4.5]{pearcecrumpC}.

\begin{theorem} \label{partfunctor}
	There exists a full, strict $\mathbb{R}$--linear monoidal functor
	\begin{equation}
		\Theta : \mathcal{P}(n) \rightarrow \mathcal{C}(S_n)
	\end{equation}
	that is defined on the objects of $\mathcal{P}(n)$ by 
	$\Theta(k) \coloneqq ((\mathbb{R}^{n})^{\otimes k}, \rho_k)$ 
	and, for any objects $k,l$ of $\mathcal{P}(n)$, the map
	\begin{equation} \label{partmorphism}
		\Hom_{\mathcal{P}(n)}(k,l) 
		\rightarrow 
		\Hom_{\mathcal{C}(S_n)}(\Theta(k),\Theta(l))
	\end{equation}
	is given by
	\begin{equation}
		d_\pi \mapsto E_\pi
	\end{equation}
	for all $(k,l)$--partition diagrams $d_\pi$,
	where $E_\pi$ is given in Theorem \ref{diagbasisSn}.
\end{theorem}

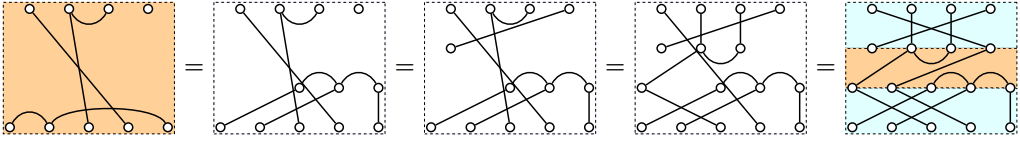
\begin{figure}
  \centering
	\scalebox{0.35}{\begin{tikzpicture}
	\begin{pgfonlayer}{nodelayer}
		\node [style=none] (0) at (14, -4.5) {\Huge$\bm{=}$};
		\node [style=node] (1) at (3, -9) {};
		\node [style=none] (2) at (-0.5, 0.5) {};
		\node [style=none] (3) at (-0.5, -9.5) {};
		\node [style=none] (4) at (12.5, -9.5) {};
		\node [style=none] (5) at (12.5, 0.5) {};
		\node [style=node] (6) at (1.5, 0) {};
		\node [style=node] (8) at (12, -9) {};
		\node [style=node] (9) at (0, -9) {};
		\node [style=node] (16) at (6, -9) {};
		\node [style=node] (17) at (9, -9) {};
		\node [style=node] (18) at (7.5, 0) {};
		\node [style=node] (19) at (4.5, 0) {};
		\node [style=node] (20) at (28, -6) {};
		\node [style=none] (21) at (15.5, 0.5) {};
		\node [style=none] (22) at (15.5, -9.5) {};
		\node [style=none] (23) at (28.5, -9.5) {};
		\node [style=none] (24) at (28.5, 0.5) {};
		\node [style=node] (25) at (17.5, 0) {};
		\node [style=node] (26) at (28, -9) {};
		\node [style=node] (27) at (25, -6) {};
		\node [style=node] (28) at (22, -9) {};
		\node [style=node] (29) at (25, -9) {};
		\node [style=node] (33) at (19, -9) {};
		\node [style=none] (37) at (30, -4.5) {\Huge$\bm{=}$};
		\node [style=node] (38) at (35, -9) {};
		\node [style=none] (39) at (31.5, 0.5) {};
		\node [style=none] (40) at (31.5, -9.5) {};
		\node [style=none] (41) at (44.5, -9.5) {};
		\node [style=none] (42) at (44.5, 0.5) {};
		\node [style=node] (43) at (33.5, 0) {};
		\node [style=node] (44) at (44, -9) {};
		\node [style=node] (45) at (32, -9) {};
		\node [style=node] (46) at (38, -9) {};
		\node [style=node] (47) at (41, -9) {};
		\node [style=node] (51) at (44, -6) {};
		\node [style=node] (52) at (41, -6) {};
		\node [style=none] (54) at (46, -4.5) {\Huge$\bm{=}$};
		\node [style=none] (56) at (47.5, 0.5) {};
		\node [style=none] (57) at (47.5, -9.5) {};
		\node [style=none] (58) at (60.5, -9.5) {};
		\node [style=none] (59) at (60.5, 0.5) {};
		\node [style=node] (60) at (49.5, 0) {};
		\node [style=node] (71) at (58.5, 0) {};
		\node [style=node] (72) at (49.5, -3) {};
		\node [style=node] (73) at (39.5, 0) {};
		\node [style=node] (74) at (36.5, 0) {};
		\node [style=node] (75) at (23.5, 0) {};
		\node [style=node] (76) at (20.5, 0) {};
		\node [style=node] (77) at (52.5, 0) {};
		\node [style=node] (78) at (55.5, 0) {};
		\node [style=none] (79) at (62, -4.5) {\Huge$\bm{=}$};
		\node [style=node] (80) at (67, -9) {};
		\node [style=none] (82) at (63.5, -9.5) {};
		\node [style=none] (83) at (76.5, -9.5) {};
		\node [style=node] (85) at (65.5, 0) {};
		\node [style=node] (86) at (76, -9) {};
		\node [style=node] (87) at (64, -9) {};
		\node [style=node] (88) at (70, -9) {};
		\node [style=node] (89) at (73, -9) {};
		\node [style=node] (90) at (64, -6) {};
		\node [style=node] (91) at (76, -6) {};
		\node [style=node] (92) at (73, -6) {};
		\node [style=node] (93) at (70, -6) {};
		\node [style=node] (94) at (71.5, -3) {};
		\node [style=node] (95) at (68.5, -3) {};
		\node [style=node] (96) at (68.5, 0) {};
		\node [style=node] (97) at (71.5, 0) {};
		\node [style=node] (98) at (65.5, -3) {};
		\node [style=node] (99) at (67, -6) {};
		\node [style=none] (102) at (63.5, -6) {};
		\node [style=none] (103) at (76.5, -6) {};
		\node [style=none] (104) at (63.5, -3) {};
		\node [style=none] (105) at (63.5, -6) {};
		\node [style=none] (106) at (76.5, -6) {};
		\node [style=none] (107) at (76.5, -3) {};
		\node [style=none] (108) at (63.5, -3) {};
		\node [style=none] (109) at (76.5, -3) {};
		\node [style=none] (110) at (76.5, 0.5) {};
		\node [style=none] (111) at (63.5, 0.5) {};
		\node [style=node] (112) at (51, -9) {};
		\node [style=node] (113) at (60, -9) {};
		\node [style=node] (114) at (48, -9) {};
		\node [style=node] (115) at (54, -9) {};
		\node [style=node] (116) at (57, -9) {};
		\node [style=node] (118) at (60, -6) {};
		\node [style=node] (119) at (57, -6) {};
		\node [style=node] (120) at (10.5, 0) {};
		\node [style=node] (121) at (26.5, 0) {};
		\node [style=node] (122) at (16, -9) {};
		\node [style=node] (123) at (42.5, 0) {};
		\node [style=node] (124) at (33.5, -3) {};
		\node [style=node] (125) at (55.5, -3) {};
		\node [style=node] (126) at (52.5, -3) {};
		\node [style=node] (127) at (48, -6) {};
		\node [style=node] (128) at (74.5, 0) {};
		\node [style=node] (129) at (74.5, -3) {};
		\node [style=node] (130) at (54, -6) {};
		\node [style=node] (131) at (38, -6) {};
		\node [style=node] (132) at (22, -6) {};
	\end{pgfonlayer}
	\begin{pgfonlayer}{edgelayer}
		\draw [style={dash_2}] (22.center) to (23.center);
		\draw [style={dash_2}] (23.center) to (24.center);
		\draw [style={dash_2}] (24.center) to (21.center);
		\draw [style={dash_2}] (21.center) to (22.center);
		\draw [style=thick] (29) to (25);
		\draw [style={dash_2}] (40.center) to (41.center);
		\draw [style={dash_2}] (41.center) to (42.center);
		\draw [style={dash_2}] (42.center) to (39.center);
		\draw [style={dash_2}] (39.center) to (40.center);
		\draw [style=thick] (47) to (43);
		\draw [style={dash_2}] (59.center) to (56.center);
		\draw [style={dash_2}] (56.center) to (57.center);
		\draw [style={dash_2}] (57.center) to (58.center);
		\draw [style={dash_2}] (58.center) to (59.center);
		\draw [style=thick, bend right=60, looseness=1.25] (74) to (73);
		\draw [style=thick, bend right=60, looseness=1.25] (76) to (75);
		\draw [style={dash_5}] (110.center)
			 to (111.center)
			 to (108.center)
			 to (109.center)
			 to cycle;
		\draw [style={dash_5}] (103.center)
			 to (102.center)
			 to (82.center)
			 to (83.center)
			 to cycle;
		\draw [style={dash_4}] (4.center)
			 to (5.center)
			 to (2.center)
			 to (3.center)
			 to cycle;
		\draw [style=thick, bend right=60, looseness=1.25] (19) to (18);
		\draw [style=thick, bend left=300, looseness=1.25] (1) to (9);
		\draw [style=thick, bend left=60, looseness=1.25] (27) to (20);
		\draw [style=thick, bend left=60, looseness=1.25] (52) to (51);
		\draw [style=thick, bend left=60, looseness=1.25] (119) to (118);
		\draw [style=thick] (60) to (116);
		\draw [style=thick] (6) to (17);
		\draw [style={dash_4}] (107.center)
			 to (104.center)
			 to (105.center)
			 to (106.center)
			 to cycle;
		\draw [style=thick, bend left=60, looseness=1.25] (92) to (91);
		\draw [style=thick, bend right=60, looseness=1.25] (95) to (94);
		\draw [style=thick, bend right=60, looseness=1.25] (126) to (125);
		\draw [style=thick] (19) to (16);
		\draw [style=thick] (76) to (28);
		\draw [style=thick] (124) to (123);
		\draw [style=thick] (74) to (46);
		\draw [style=thick] (126) to (77);
		\draw [style=thick] (78) to (125);
		\draw [style=thick] (72) to (71);
		\draw [style=thick] (127) to (126);
		\draw [style=thick] (115) to (127);
		\draw [style=thick] (90) to (95);
		\draw [style=thick] (95) to (96);
		\draw [style=thick] (94) to (97);
		\draw [style=thick] (129) to (85);
		\draw [style=thick] (98) to (128);
		\draw [style=thick] (88) to (90);
		\draw [style=thick] (89) to (99);
		\draw [style=thick] (99) to (129);
		\draw [style=thick, bend left=300, looseness=1.25] (92) to (93);
		\draw [style=thick] (87) to (93);
		\draw [style=thick] (80) to (92);
		\draw [style=thick] (86) to (91);
		\draw [style=thick] (114) to (130);
		\draw [style=thick] (112) to (119);
		\draw [style=thick] (113) to (118);
		\draw [style=thick, bend left=60, looseness=1.25] (130) to (119);
		\draw [style=thick, bend left=60, looseness=1.25] (131) to (52);
		\draw [style=thick] (44) to (51);
		\draw [style=thick] (38) to (52);
		\draw [style=thick] (45) to (131);
		\draw [style=thick, bend left=60, looseness=1.25] (132) to (27);
		\draw [style=thick] (122) to (132);
		\draw [style=thick] (33) to (27);
		\draw [style=thick] (26) to (20);
		\draw [style=thick, bend left=60, looseness=0.50] (1) to (8);
	\end{pgfonlayer}
\end{tikzpicture}}
	\caption{
	We use the string-like aspect of $(k,l)$--partition diagrams to \textsc{Factor} them as a composition of a permutation in $S_k$, a \textit{planar} $(k,l)$--partition diagram, and a permutation in $S_l$.}
	\label{symmfactoringSn}
\end{figure}

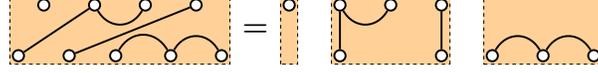
\begin{figure}
	\centering
	\scalebox{0.45}{\begin{tikzpicture}
	\begin{pgfonlayer}{nodelayer}
		\node [style=node] (0) at (0, -3) {};
		\node [style=node] (1) at (12, -3) {};
		\node [style=node] (2) at (9, -3) {};
		\node [style=node] (3) at (5.75, -3) {};
		\node [style=node] (4) at (7.5, 0) {};
		\node [style=node] (5) at (4.5, 0) {};
		\node [style=node] (6) at (1.5, 0) {};
		\node [style=node] (7) at (3, -3) {};
		\node [style=none] (10) at (-0.5, 0.5) {};
		\node [style=none] (11) at (-0.5, -3.5) {};
		\node [style=none] (12) at (12.5, -3.5) {};
		\node [style=none] (13) at (12.5, 0.5) {};
		\node [style=none] (14) at (14, -1.5) {\Huge$\bm{=}$};
		\node [style=node] (15) at (16, 0) {};
		\node [style=node] (16) at (31, -3) {};
		\node [style=node] (17) at (28, -3) {};
		\node [style=node] (18) at (25, -3) {};
		\node [style=node] (19) at (22, 0) {};
		\node [style=node] (20) at (19, 0) {};
		\node [style=node] (21) at (25, 0) {};
		\node [style=none] (23) at (15.5, -3.5) {};
		\node [style=none] (24) at (15.5, 0.5) {};
		\node [style=none] (25) at (16.5, 0.5) {};
		\node [style=none] (26) at (16.5, -3.5) {};
		\node [style=none] (27) at (34.5, -3.5) {};
		\node [style=none] (28) at (27.5, -3.5) {};
		\node [style=none] (29) at (34.5, 0.5) {};
		\node [style=none] (30) at (27.5, 0.5) {};
		\node [style=none] (35) at (25.5, -3.5) {};
		\node [style=none] (36) at (18.5, -3.5) {};
		\node [style=none] (37) at (25.5, 0.5) {};
		\node [style=none] (38) at (18.5, 0.5) {};
		\node [style=node] (39) at (10.5, 0) {};
		\node [style=node] (40) at (19, -3) {};
		\node [style=node] (41) at (34, -3) {};
	\end{pgfonlayer}
	\begin{pgfonlayer}{edgelayer}
		\draw [style={dash_4}] (13.center)
			 to (10.center)
			 to (11.center)
			 to (12.center)
			 to cycle;
		\draw [style=thick, bend left=60, looseness=1.25] (2) to (1);
		\draw [style=thick, bend right=60, looseness=1.25] (5) to (4);
		\draw [style={dash_4}] (23.center)
			 to (24.center)
			 to (25.center)
			 to (26.center)
			 to cycle;
		\draw [style={dash_4}] (35.center)
			 to (37.center)
			 to (38.center)
			 to (36.center)
			 to cycle;
		\draw [style={dash_4}] (28.center)
			 to (27.center)
			 to (29.center)
			 to (30.center)
			 to cycle;
		\draw [style=thick, bend left=60, looseness=1.25] (17) to (16);
		\draw [style=thick] (18) to (21);
		\draw [style=thick, bend right=60, looseness=1.25] (20) to (19);
		\draw [style=thick] (0) to (5);
		\draw [style=thick] (40) to (20);
		\draw [style=thick] (7) to (39);
		\draw [style=thick, bend left=60, looseness=1.25] (3) to (2);
		\draw [style=thick, bend left=60, looseness=1.25] (16) to (41);
	\end{pgfonlayer}
\end{tikzpicture}}
	\caption{The decomposition of the planar $(5,4)$--partition diagram that appears in Figure \ref{symmfactoringSn} into a tensor product of smaller partition diagrams. These diagrams correspond under the functor $\Theta$ to operations that perform, from right--to--left, tensor contraction, transferring, and copying, respectively, on an input vector.
	This tensor product decomposition is used in \textsc{PlanarMult} for the symmetric group $S_n$.}
	\label{tensorproddecompSn}
\end{figure}

To implement the algorithm, it is enough to consider how
$v \in (\mathbb{R}^{n})^{\otimes k}$ is mapped
by a basis matrix
$E_\pi \in \Hom_{S_n}((\mathbb{R}^{n})^{\otimes k}, (\mathbb{R}^{n})^{\otimes l})$,
where $d_\pi$ is a $(k,l)$--partition diagram.

As before, we provide new implementations of \textsc{Factor} and \textsc{PlanarMult}.

\textsc{Factor}: The input is $d_\pi$. 
We drag and bend the strings representing the connected components of $d_\pi$ to obtain a factoring into three diagrams whose composition is equivalent to $d_\pi$:
a $(k,k)$--partition diagram that represents a permutation $\sigma_k$ in the symmetric group $S_k$; another $(k,l)$--partition diagram that is planar; and a $(l,l)$--partition diagram that represents a permutation $\sigma_l$ in the symmetric group $S_l$.
To obtain the desired planar $(k,l)$--partition diagram, 
we drag and bend the strings in any way such that 
\begin{itemize}
	\item the connected components that are solely in the bottom row of $d_\pi$ are pulled up to be next to each other
		in the far right hand side of the bottom row of the planar $(k,l)$--partition diagram,
	\item the connected components that are solely in the top row of $d_\pi$ are pulled down to be next to each other
		in the far left hand side of the top row of the planar $(k,l)$--partition diagram, and
	\item the connected components between vertices in different rows of $d_\pi$ are bent to be in between the other vertices of the planar $(k,l)$--partition diagram such that no two subsets of connected components cross.
\end{itemize}
We give an example of this procedure in Figure \ref{symmfactoringSn}.

\textsc{PlanarMult}: we take the planar $(k,l)$--partition diagram that comes from \textsc{Factor} and express it as a tensor product of three types of set partition diagrams.
The right-most type is itself a tensor product of set partition diagrams having only vertices in the bottom row, where each diagram represents a single block in the planar partition.
These diagrams correspond to tensor contraction operations under the functor $\Theta$.
The middle type is a set partition diagram that consists of all of the connected components in the planar $(k,l)$--partition diagram between vertices in different rows.
These diagrams correspond to transfer operations under the functor $\Theta$.
The left-most type is itself a tensor product of set partition diagrams having only vertices in the top row, where each diagram represents a single block.
These diagrams correspond to indexing operations that perform copies under the functor $\Theta$.

Figure \ref{tensorproddecompSn} gives an example of the tensor product decomposition for the planar $(5,4)$--partition diagram given in Figure \ref{symmfactoringSn}.

\begin{figure}
  		\centering
		\scalebox{0.35}{\input{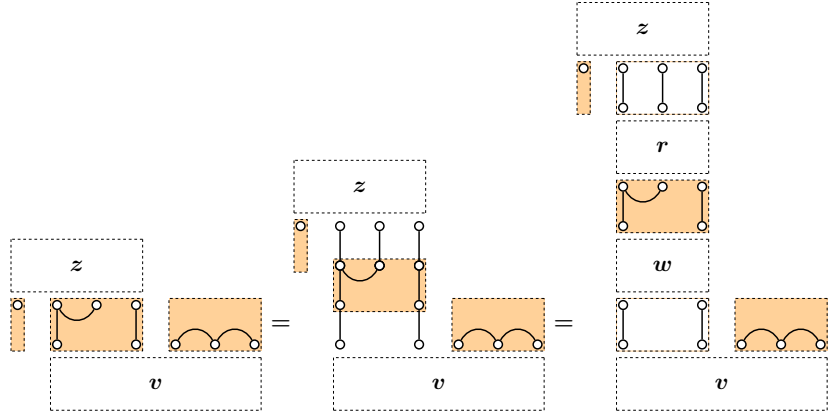}}
		\caption{We show how matrix multiplication is implemented in \textsc{PlanarMult} for $S_n$ using the tensor product decomposition of the planar 
		$(5,4)$--partition diagram
		given in Figure \ref{tensorproddecompSn}
		as an example.
		We perform the matrix multiplication as follows:
		first, we deform the entire tensor product decomposition diagram by pulling each individual diagram up one level higher than the previous one, going from right--to-left, and then we apply the functor $\Theta$ at each level. 
		Finally, we perform matrix multiplication at each level to obtain the final output vector.}
		\label{planarmultSn}
	\end{figure}

The matrix multiplication step is very similar to the orthogonal group, in that to obtain the matrices we perform the same deformation of the tensor product decomposition of diagrams before applying the functor $\Theta$ given in Theorem \ref{partfunctor}
at each level.
We give an example in Figure \ref{planarmultSn}
of how the computation would take place at each stage, using its equivalent diagram form.


We now describe in full how the matrix multiplication is performed in the \textsc{PlanarMult} procedure for $S_n$.

	We take as input the planar $(k,l)$--partition diagram $d_\pi$ having at most $n$ blocks that is the output of \textsc{Factor}, and a vector $v \in (\mathbb{R}^{n})^{\otimes k}$ that is the output of \textsc{Permute}, as per Algorithm \ref{alg1}.
	
	Let $t$ be the number of blocks that are solely in the top row of $d_\pi$,
	let $d$ be the number of blocks that are connect vertices in different rows of $d_\pi$, and
	let $b$ be the number of blocks that are solely in the bottom row of $d_\pi$.


	Given how \textsc{Factor} constructs the planar $(k,l)$--partition diagram $d_\pi$, the set partition $\pi$ corresponding to $d_\pi$ will be of the form
	\begin{equation} \label{symmpartfactor}
		\left(\bigcup_{i = 1}^{t} T_i \right)
		\bigcup
		\left(\bigcup_{i = 1}^{d} D_i \right)
		\bigcup
		\left(\bigcup_{i = 1}^{b} B_i \right)
	\end{equation}
	where we have used $T_i$ to refer to a top row block, $D_i$ to refer to a different row block, and $B_i$ to refer to a bottom row block.
	Defining
	\begin{equation}
		D_i \coloneqq D_i^{U} \cup D_i^{L}
	\end{equation}
	where $D_i^{U}$ is the subset of $D_i$ whose vertices are in the top row of $d_\pi$, and
	$D_i^{L}$ is the subset of $D_i$ whose vertices are in the bottom row of $d_\pi$,
	we have that
	\begin{itemize}
		\item $T_i = \{
				\sum_{j=1}^{i-1} |T_j| + 1, \dots, \sum_{j=1}^{i} |T_j|
			\}$ for all $i = 1 \rightarrow t$
		\item $D_i^{U} = \{
				\sum_{j=1}^{t} |T_j| + \sum_{j=1}^{i-1} |D_j^{U}| + 1, \dots, 
		\sum_{j=1}^{t} |T_j| + \sum_{j=1}^{i} |D_j^{U}|
			\}$ for all $i = 1 \rightarrow d$,
		\item $D_i^{L} = \{
				l + \sum_{j=1}^{i-1} |D_j^{L}| + 1, \dots, 
			l + \sum_{j=1}^{i} |D_j^{L}|
			\}$ for all $i = 1 \rightarrow d$, and
		\item $B_i = \{
			l + \sum_{j=1}^{d} |D_j^{L}| + 
			\sum_{j=1}^{i-1} |B_j| + 1, \dots, 
			l + \sum_{j=1}^{d} |D_j^{L}| +
			\sum_{j=1}^{i} |B_j|
			\}$ for all $i = 1 \rightarrow b$.
	\end{itemize}
	Note, in particular, that
	\begin{equation} \label{symmupper}
		l = \sum_{j = 1}^{t} |T_j| + \sum_{j = 1}^{d} |D_j^{U}|
	\end{equation}
	and
	\begin{equation} \label{symmlower}
		k = \sum_{j = 1}^{d} |D_j^{L}| + \sum_{j = 1}^{b} |B_j|
	\end{equation}
	Next, we take $d_\pi$ and express it as a tensor product of three types of set partition diagrams.
	The right-most type is itself a tensor product over diagrams corresponding to the $B_i$; the middle type is a single diagram corresponding to $\bigcup_{i = 1}^{d} D_i$; and the left-most type is itself a tensor product over diagrams corresponding to the $T_i$.

	We now apply the monoidal functor $\Theta$ to this tensor product decomposition of diagrams, which returns a Kronecker product of matrices.
	We perform the matrix multiplication by applying the matrices "right--to--left, diagram--by--diagram", as follows.

	\textbf{Step 1: Apply each matrix corresponding to a bottom row pair diagram, one by one, starting from the one that corresponds to $B_b$ and ending with the one that corresponds to $B_1$.}

	Suppose that we are performing the part of the matrix multiplication that corresponds to $B_i$, for some $i = 1 \rightarrow b$.
	The input will be a vector
	$w \in (\mathbb{R}^{n})^{\otimes k - \sum_{j = i+1}^{b} |B_j|}$.

	We can express $w$ in the standard basis of $\mathbb{R}^{n}$ as
	\begin{equation}
		w = \sum_{L \in [n]^{k - \sum_{j = i+1}^{b} |B_j|}}
			w_Le_L
	\end{equation}

	This will be mapped to the vector
	$r \in (\mathbb{R}^{n})^{\otimes k - \sum_{j = i}^{b} |B_j|}$,
	where $r$ is of the form
	\begin{equation}
		r = \sum_{M \in [n]^{k - \sum_{j = i}^{b} |B_j|}} r_Me_M
	\end{equation}
	and
	\begin{equation} \label{indicesj}
		r_M = \sum_{j=1}^{n} w_{M,j, \dots, j}
	\end{equation}
	where the number of indices $j$ in (\ref{indicesj}) is $|B_i|$.

	At the end of this process, we obtain a vector in 
	$(\mathbb{R}^{n})^{\otimes k - \sum_{j = 1}^{b} |B_j|}$.
	
	Note that the matrices corresponding to bottom row pairs are merely performing indexing and summation operations, that is, ultimately, tensor contractions.

	\textbf{Step 2: Now apply the matrix corresponding to the middle diagram, that is, to the set $\bigcup_{i = 1}^{d} D_i$.}
	
	The input will be a vector
	$w \in 
	(\mathbb{R}^{n})^{\otimes k - \sum_{j = 1}^{b} |B_j|}$.
	Using (\ref{symmlower}), we can express $w$ in the standard basis of $\mathbb{R}^{n}$ as
	\begin{equation}
		w = 
		\sum_{j_1, \dots, j_d \in [n]} 
		w_{j_1, \dots, j_1, j_2, \dots, j_2, \dots, j_d, \dots, j_d}
		\bigotimes_{k = 1}^{d}
		\left(
		\bigotimes_{n = 1}^{|D_k^{L}|} e_{j_k}
		\right)
	\end{equation}
	where each index $j_k$ in the coefficient is repeated $|D_k^{L}|$ times.
	
	This will be mapped to the vector 
	$r \in (\mathbb{R}^{n})^{\otimes \sum_{j = i}^{d} |D_j^{U}|}$,
	where $r$ is of the form
	\begin{equation}
		r = 
		\sum_{j_1, \dots, j_d \in [n]} 
		r_{j_1, \dots, j_1, j_2, \dots, j_2, \dots, j_d, \dots, j_d}
		\bigotimes_{k = 1}^{d}
		\left(
		\bigotimes_{n = 1}^{|D_k^{U}|} e_{j_k}
		\right)
	\end{equation}
	where each index $j_k$ in the coefficient is repeated $|D_k^{U}|$ times, and
	\begin{equation} \label{transfercoeffSn}
		r_{j_1, \dots, j_1, j_2, \dots, j_2, \dots, j_d, \dots, j_d}
		=
		w_{j_1, \dots, j_1, j_2, \dots, j_2, \dots, j_d, \dots, j_d}
	\end{equation}
	These operations are called transfer operations, as first described in \cite{pan22}.

	\textbf{Step 3: Finally, apply each matrix corresponding to a top row pair diagram, one by one, starting from the one that corresponds to $T_t$ and ending with the one that corresponds to $T_1$.}
	
	Suppose that we are performing the part of the matrix multiplication that corresponds to $T_i$, for some $i = 1 \rightarrow t$.

	Then we begin with a vector  
	$x \in (\mathbb{R}^{n})^{\otimes l - \sum_{j=1}^{i} |T_j|}$
	that is of the form
	\begin{equation}
		x = 
		\sum_{l_{i+1}, \dots, l_{t} \in [n]}
		\sum_{j_1, \dots, j_d \in [n]} 
		x_{j_1, j_2, \dots, j_d}
		\bigotimes_{q = i+1}^{t}
		\left(
		\bigotimes_{m = 1}^{|T_q|} e_{l_q}
		\right)
		\bigotimes_{k = 1}^{d}
		\left(
		\bigotimes_{n = 1}^{|D_k^{U}|} e_{j_k}
		\right)
	\end{equation}
	where $x_{j_1, j_2, \dots, j_d}$ is the coefficient 
	$r_{j_1, \dots, j_1, j_2, \dots, j_2, \dots, j_d, \dots, j_d}$
	appearing in (\ref{transfercoeffSn}).
	
	This will be mapped to the vector
	$y \in (\mathbb{R}^{n})^{\otimes l - \sum_{j=1}^{i-1} |T_j|}$,
	where $y$ is of the form
	\begin{equation}
		y = 
		\sum_{l_i, l_{i+1}, \dots, l_{t} \in [n]}
		\sum_{j_1, \dots, j_d \in [n]} 
		x_{j_1, j_2, \dots, j_d}
		\bigotimes_{q = i}^{t}
		\left(
		\bigotimes_{m = 1}^{|T_q|} e_{l_q}
		\right)
		\bigotimes_{k = 1}^{d}
		\left(
		\bigotimes_{n = 1}^{|D_k^{U}|} e_{j_k}
		\right)
	\end{equation}
	At the end of this process, we obtain a vector in 
	$(\mathbb{R}^{n})^{\otimes l}$, by (\ref{symmupper}),
	which is of the form
	\begin{equation}
		\sum_{l_1, \dots, l_{t} \in [n]}
		\sum_{j_1, \dots, j_d \in [n]} 
		x_{j_1, j_2, \dots, j_d}
		\bigotimes_{q = 1}^{t}
		\left(
		\bigotimes_{m = 1}^{|T_q|} e_{l_q}
		\right)
		\bigotimes_{k = 1}^{d}
		\left(
		\bigotimes_{n = 1}^{|D_k^{U}|} e_{j_k}
		\right)
	\end{equation}
	This is the vector that is returned by \textsc{PlanarMult} for the symmetric group $S_n$.

	We give an example in Appendix \ref{MatrixMultExamplesSn}.

\begin{remark}
	The implementation of Algorithm \ref{alg1} that we have given here for the symmetric group effectively recovers the one given in \cite[Appendix C]{godfrey}; however, we have used an entirely different approach -- involving monoidal categories -- to obtain it.
	The major difference in implementation between the two versions relates to how (in our terminology)
	the connected components between vertices in different rows of $d_\pi$ are pulled into the middle diagram in \textsc{Factor}.
	We choose to make sure that these do not cross, thus making the resulting diagram \textit{planar}, whereas in \cite{godfrey} they choose to connect the vertices in different rows such that the left-most vertices in the top row of the new diagram connect to the right-most vertices in the bottom row, thus making the components cross in "opposites".
	Whilst this doesn't make any significant difference in terms of performing the matrix multiplication for the symmetric group (there is only a difference in how the tensor indices are ordered), our decision to make the middle diagram in the composition planar leads to significant performance improvements when we extend our approach to the other groups in the main text, since for these groups, these operations reduce to the identity transformation!
	See \cite[Appendix C]{godfrey} for a computational cost analysis of the algorithm.
\end{remark}

	\subsubsection{Example of \textsc{MatrixMult} for the Symmetric Group $S_n$}
	\label{MatrixMultExamplesSn}

	Suppose that we wish to perform the multiplication of $E_\pi$
	by $v \in (\mathbb{R}^{n})^{\otimes 5}$, where
	$E_\pi$ corresponds to the $(5,4)$--partition diagram $d_\pi$ given in Figure \ref{symmfactoringSn},
	and $v$ is given by
		\begin{equation}
			\sum_{L \in [n]^5} v_Le_L
		\end{equation}
	Here, we assume that $n \geq 4$, since the number of blocks in $\pi$ is $4$.
	We know that $E_\pi$ is a matrix in
	$\Hom_{S_n}((\mathbb{R}^{n})^{\otimes 5}, (\mathbb{R}^{n})^{\otimes 4})$.

	First, we apply the procedure \textsc{Factor}, which returns the three diagrams given in Figure \ref{symmfactoringSn}.
	The first diagram corresponds to the permutation $(13)(24)$ in $S_5$, hence, the result of \textsc{Permute}$(v, (13)(24))$ is the vector
		\begin{equation}
			\sum_{L \in [n]^5} 
			v_{l_1, l_2, l_3, l_4, l_5}
			\left(
				e_{l_3} \otimes 
				e_{l_4} \otimes 
				e_{l_1} \otimes
				e_{l_2} \otimes
				e_{l_5} 
			\right)
		\end{equation}
	Now we apply \textsc{PlanarMult} with the decomposition given in Figure 
\ref{tensorproddecompSn}.

	Step 1: Apply the matrices that correspond to the bottom row blocks.

	We obtain the vector
		\begin{equation}
			w = 
			\sum_{l_3, l_4 \in [n]} 
			w_{l_3, l_4}
			\left(
				e_{l_3} \otimes 
				e_{l_4}
			\right)
		\end{equation}
	where
		\begin{equation}
			w_{l_3, l_4}
			=
			\sum_{j \in [n]}
			v_{j, j, l_3, l_4, j}
		\end{equation}

	Step 2: Apply the matrices that correspond to the middle diagram.

	We obtain the vector
		\begin{equation}
			r = 
			\sum_{l_3 \in [n]}
			\sum_{l_4 \in [n]}
			r_{l_3, l_3, l_4}
			\left(	
				e_{l_3} \otimes 
				e_{l_3} \otimes 
				e_{l_4}
			\right)
		\end{equation}
	where
		\begin{equation}
			r_{l_3, l_3, l_4}
			=
			w_{l_3, l_4}
		\end{equation}	
	Step 3: Apply the matrices that correspond to the top row blocks.

	We obtain the vector
		\begin{equation}
			z =
			\sum_{m \in [n]}	
			\sum_{l_3 \in [n]}
			\sum_{l_4 \in [n]}
			r_{l_3, l_3, l_4}
			\left(
				e_{m} \otimes 
				e_{l_3} \otimes 
				e_{l_3} \otimes 
				e_{l_4}
			\right)
		\end{equation}
	Substituting in, we get that
		\begin{equation}
			z =
			\sum_{m \in [n]}	
			\sum_{l_3 \in [n]}
			\sum_{l_4 \in [n]}
			w_{l_3, l_4}
			\left(
				e_{m} \otimes 
				e_{l_3} \otimes 
				e_{l_3} \otimes 
				e_{l_4}
			\right)
		\end{equation}
	and hence
		\begin{equation}
			z =
			\sum_{m \in [n]}	
			\sum_{l_3 \in [n]}
			\sum_{l_4 \in [n]}
			\sum_{j \in [n]}
			v_{j, j, l_3, l_4, j}
			\left(
				e_{m} \otimes 
				e_{l_3} \otimes 
				e_{l_3} \otimes 
				e_{l_4}
			\right)
		\end{equation}
	Finally, as the third diagram returned from \textsc{Factor} 
	corresponds to the permutation $(14)$ in $S_4$,
	we perform \textsc{Permute}$(z, (14))$,
	which returns the vector
		\begin{equation}
			z =
			\sum_{m \in [n]}	
			\sum_{l_3 \in [n]}
			\sum_{l_4 \in [n]}
			\sum_{j \in [n]}
			v_{j, j, l_3, l_4, j}
			\left(
				e_{l_4} \otimes 
				e_{l_3} \otimes 
				e_{l_3} \otimes 
				e_{m}
			\right)
		\end{equation}
	This is the vector that is returned by \textsc{MatrixMult}.

\end{appendix}

\end{document}